\documentclass[final,5p,times,twocolumn,authoryear]{elsarticle}

\usepackage{amssymb}
\usepackage{soul}
\usepackage{booktabs}
\usepackage{multirow}
\usepackage{tikz}
\usepackage{colortbl}
\usepackage{tcolorbox}
\usepackage{algorithm}
\usepackage{algorithmic}
\usepackage{array}
\usepackage{amsmath}
\usepackage{hyperref}
\usepackage{dsfont} 
\usepackage{enumitem}
\usepackage{xcolor}
\usepackage{makecell}
\journal{Results in Engineering}

\makeatletter
\def\ps@pprintTitle{%
    \let\@oddhead\@empty
    \let\@evenhead\@empty
    \let\@oddfoot\@empty
    \let\@evenfoot\@empty
}
\makeatother

\begin{document}

\begin{frontmatter}



\title{On the Black-box Explainability of Object Detection Models for\\Safe and Trustworthy Industrial Applications\tnoteref{label1}}

\tnotetext[label1]{This paper has been accepted for publication in \textbf{Results in Engineering}, DOI: \url{https://doi.org/10.1016/j.rineng.2024.103498}.}

\author[tecnalia,deusto]{Alain Andres\corref{cor1}}
\author[tecnalia]{Aitor Martinez-Seras}
\author[tecnalia,deusto]{Ibai La\~na}
\author[tecnalia,upv]{Javier Del Ser}

\affiliation[tecnalia]{organization={TECNALIA, Basque Research and Technology Alliance (BRTA)},
            addressline={Mikeletegi Pasealekua 2}, 
            city={Donostia-San Sebastian},
            postcode={20009}, 
            country={Spain}}

\affiliation[deusto]{
        organization={University of Deusto},
        postcode={20012}, 
        city={Donostia-San Sebastián},
        country={Spain}}
        
\affiliation[upv]{organization={University of the Basque Country (UPV/EHU)},
            city={Bilbao},
            postcode={48013}, 
            country={Spain}}

\cortext[cor1]{Corresponding author: alain.andres@tecnalia.com}

\begin{abstract}
In the realm of human-machine interaction, artificial intelligence has become a powerful tool for accelerating data modeling tasks. Object detection methods have achieved outstanding results and are widely used in critical domains like autonomous driving and video surveillance. However, their adoption in high-risk applications, where errors may cause severe consequences, remains limited. Explainable Artificial Intelligence methods aim to address this issue, but many existing techniques are model-specific and designed for classification tasks, making them less effective for object detection and difficult for non-specialists to interpret. In this work we focus on \emph{model-agnostic} explainability methods for object detection models and propose D-MFPP, an extension of the Morphological Fragmental Perturbation Pyramid (MFPP) technique 
based on segmentation-based masks to generate explanations. 
Additionally, we introduce D-Deletion, a novel metric combining faithfulness and localization, adapted specifically to meet the unique demands of object detectors. We evaluate these methods on real-world industrial and robotic datasets, examining the influence of parameters such as the number of masks, model size, and image resolution on the quality of explanations. Our experiments use single-stage object detection models applied to two safety-critical robotic environments: i) a shared human-robot workspace where safety is of paramount importance, and ii) an assembly area of battery kits, where safety is critical due to the potential for damage among high-risk components. Our findings evince that D-Deletion effectively gauges the performance of explanations when multiple elements of the same class appear in 
a scene, while D-MFPP provides a promising alternative to D-RISE when fewer masks are used.

\end{abstract}

\begin{keyword}
Explainable Artificial Intelligence \sep
Safe Artificial Intelligence \sep
Trustworthy Artificial Intelligence \sep
Object Detection \sep
Single-stage object detection\sep
Industrial Robotics
\end{keyword}

\end{frontmatter}

\section{Introduction}\label{}

In recent years, Artificial Intelligence (AI) has emerged as a transformative force across various domains, especially in human-machine interaction, where it has enabled significant advancements in data-driven decision-making processes. Among these advances, object detection has become a key component, finding application in critical areas such as autonomous driving, security surveillance, industrial automation, and robotics \cite{zou2023object,muhammad2020deep}. State-of-the-art object detection models, including Faster-RCNN \cite{ren_faster_2017}, DETR \cite{vedaldi_end--end_2020}, and the YOLO series \cite{terven_comprehensive_2023}, have demonstrated impressive performance in identifying and localizing objects within images. Despite their success, the adoption of these models in highly sensitive environments remains limited, particularly in domains where errors could result in serious consequences such as injury, equipment damage, or operational failures. One of the primary reasons for this hesitancy is the black-box nature of object detectors implemented as Deep Learning models, which to date amount to the majority of proposals in the literature. The internal activations of these are not inherently interpretable, making it challenging for end-users to trust the predictions issued by object detectors, especially in high-risk environments operating in open-world environments such as autonomous vehicles and industrial robotics. 

In this context, the field of Explainable AI (XAI) \cite{barredo_arrieta_explainable_2020} aims to enhance the interpretability of AI systems by their \emph{audience} and ultimately, to enhance the user's trust in the output of AI-based systems. Leaving aside the category of transparent AI models (which are inherently interpretable and do not require any explanations for a user to understand how they work), explainability methods in XAI can be broadly categorized into \emph{white-box} and \emph{black-box} approaches. \emph{White-box XAI methods} require access to the internal workings of the model, such as weights, activations, or gradients
(e.g., Grad-CAM \cite{selvaraju_grad-cam_2017}). While these methods can provide powerful insights, they are often limited by their dependence on specific model architectures, making them difficult to generalize across different models and less accessible to users unfamiliar with AI research/tools. In contrast, \emph{black-box XAI methods} treat the model as an opaque entity, providing explanations based solely on the model's input-output behavior without requiring any access to its internal components. However, most black-box XAI methods are designed for classification tasks rather than for object detection \cite{ribeiro_why_2016,lundberg_unified_2017,petsiuk_rise_2018, ali_explainable_2023}. 

While classification models produce a single label per image, object detection models must identify and localize multiple objects within an image. Therefore, they need to explain not only the class prediction for each detected object -- \emph{what} they detect-- but also the spatial reasoning behind the bounding boxes that define the object’s location -- \emph{where} the object is positioned within the image. Balancing these dual aspects complicates the explanation process and requires more sophisticated techniques than those used for classification tasks.

In this paper, we address the gap in XAI methods for object detection by focusing on \emph{model-agnostic}, \emph{black-box} XAI techniques. We propose and evaluate novel black-box XAI methods and XAI metrics that are specifically tailored for object detection models, without requiring access to internal model details. Our proposed methods are generalizable to object detection frameworks beyond those utilized in our experiments. Specifically, the contributions of this work can be summarized as follows:
\begin{itemize}[leftmargin=*]
    \item We formally define a quantitative evaluation metric, \textbf{D-Deletion}, which extends the existing Deletion metric \cite{bach_pixel-wise_2015,petsiuk_rise_2018} proposed for classification tasks. This metric is adapted to handle the unique challenges of object detection, including localization (as seen in Figure \ref{fig:d-deletion_motivation}), which is of utmost importance when multiple instances of the same object appear in the same scene.
    \item By using the similarity score of D-RISE \cite{petsiuk_black-box_2021}, we analyze multiple mask generation methods' performance and introduce \textbf{D-MFPP}, an extension of MFPP \cite{yang_mfpp_2020} originally developed for classification tasks. D-MFPP utilizes segmentation-based mask generation to improve explanations for object detection models.
    \item We analyze the impact of key parameters, such as image dimensions and the model sizes within the YOLOv8 architecture utilized in our experiments, which can significantly influence the quality of the resulting explanations.
    \item Last but not least, we facilitate the broader adoption of the developed techniques for object detection in real-world use cases by releasing the code publicly in a repository: 
    \url{https://github.com/aklein1995/drise_dmfpp_ddeletion}.
\end{itemize}

The remainder of this paper is structured as follows: in Section \ref{sec:related_work}, we first review literature related to XAI for object detection. 
In Section \ref{sec:background}, we provide the necessary background on object detection and XAI to familiarize the reader with the key concepts used in the definitions of D-RISE and Deletion. Next, Section \ref{sec:materials_and_methods} presents the experimental setup, including datasets, object detection training configuration, employed XAI methods, and evaluation metrics. In this section we also introduce our proposed D-MFPP method and D-Deletion metric.
We discuss our results in Section \ref{sec:results}. Finally, Section \ref{sec:conclusions} concludes the paper with a summary of our key findings and directions for future research.

\section{Related Work}\label{sec:related_work}

Before proceeding with the materials and novel methods introduced in this work, we first pause briefly at XAI methods, focusing on those used for object detection tasks and put to practice in industrial applications:

\paragraph{XAI methods}As stated in the introduction, XAI offers insights into the procedure followed by an AI-based system to elicit their outputs, enabling end-users to understand and eventually trust the decisions output by the AI-based system grounded on objective data \cite{ali_explainable_2023}. To date, the majority of XAI methods are designed for models learned to address classification tasks. For instance, CAM-based methods like GradCAM \cite{selvaraju_grad-cam_2017}, GradCAM++\cite{chattopadhay_grad-cam_2018} and Integrated Gradients \cite{sundararajan_axiomatic_2017} quantify and attribute the pixel-wise importance of a given input according to the gradients with respect a target class. Moreover, making use of backpropagation, LRP \cite{montavon_layer-wise_2019} calculates the contribution that a neuron has with neurons in consecutive layers to get relevance scores. In contrast, perturbation-based techniques work by occluding certain parts of the input and analyzing its impact in the predictions. Within this type of techniques, LIME \cite{ribeiro_why_2016}, approximates a NN with an interpretable model; SHAP \cite{lundberg_unified_2017} assigns importance values to each input feature based on Shapley values; RISE \cite{petsiuk_rise_2018} generates saliency maps by probing the model with randomly masked versions of the input image; and MFPP \cite{yang_mfpp_2020} generates masks by dividing the input image into multi-scale superpixels. 
Nonetheless, none of them have been explicitly extended for object detection tasks --with the exception of RISE, which has been adapted for this purpose-- although techniques like SHAP can also be utilized for regression problems.

\paragraph{XAI methods for object detection models}In recent times, a scarcity of XAI approaches has been proposed to support the interpretability of complex object detection models. SODEx \cite{sejr_surrogate_2021}
is a method capable of explaining any object detection algorithm using classification explainers, demonstrating how LIME can be integrated within YOLOv4, a variant of the YOLO family of single-stage object detectors. Similarly, D-RISE \cite{petsiuk_black-box_2021} extends RISE's mask generation technique by introducing a new similarity score that assesses both the localization and classification aspects of object detection models.
More recently, D-CLOSE \cite{truong_towards_2024} enhances D-RISE by producing less noisy explanations. Along with other methodological improvements, D-CLOSE uses multiple levels of segmentation in the mask generation phase. Other approaches focusing on hierarchical masking have been proposed. Concretely, GSM-NH \cite{yan_gsm-hm_2022} evaluates the saliency maps at multiple levels based on the information of previous less fine-grained saliency maps, whereas BODEM \cite{moradi_model-agnostic_2024} further extends this idea but focuses on an extreme black-box scenario where only object coordinates are available.

\vspace{1mm}
\paragraph{XAI methods for industrial applications}Although XAI is increasingly important in industrial settings to ensure safety, reliability and compliance, the adoption of XAI for object detection methods in industrial use cases has been limited to date \cite{le_exploring_2023,kotriwala_xai_2021,chen_explainable_2023}.
The vast majority of the works focus either on image classification, like \cite{chen_vibration_2020} that utilizes Grad-CAM to interpret vibration signal images in the classification of bearing faults; time-series data, e.g. \cite{serradilla_interpreting_2020} that presents the implementation and explanations of a remaining life estimator model;
or tabular data, as in \cite{ryo_explainable_2022} where SHAP is used to interpret and study the influence of soil and climate features on crop recommendations. Regarding XAI and object detection for industrial applications, we can find a few exemplary studies that expose the shortage of real-world use cases currently noted in this technological crossroads.  
In \cite{naddaf-sh_real-time_2022}, various object detection models are evaluated for their effectiveness in detecting weld characteristics in radiography images, with an emphasis on explainability and deployment on edge devices to assist workers. In the same sense, \cite{sahatova_overview_2022} provides a comprehensive review and analysis of various XAI techniques applied to object detection tasks in computerized tomography imaging for medical purposes. Finally, \cite{kirchknopf_explaining_2022} demonstrates how to integrate Grad-CAM into the YOLO architecture and performs experiments in both public and private datasets of vehicle front collision and rear-view cameras.

\section{Background}\label{sec:background}

We now proceed by elaborating on key concepts needed to properly understand the details of the proposed D-MFPP technique and the D-Deletion metric that lie at the core of this work. Concretely, we provide fundamentals for object detection models (Section \ref{backgroud:objdet}) 
and XAI, with a focus on model-agnostic black-box methods to explain the predictions of object detection models (Section \ref{backgroud:xai}).

\subsection{Object Detectors}\label{backgroud:objdet}

Object detectors are crucial components in computer vision tasks, capable of identifying and localizing objects within an image. They can be broadly categorized into single-stage and two-stage detectors.

\paragraph{Single-stage detectors}
They directly predict bounding boxes and class probabilities from input images in a single pass. 
Popular single-stage detectors, such as YOLO \cite{terven_comprehensive_2023}, SSD \cite{liu2016ssd} and RetinaNet \cite{ross2017focal},
treat object detection as a simple regression problem, straight from image pixels to bounding box coordinates and class probabilities. To this end, they produce a dense grid of bounding box proposals and class probabilities in one step. Specifically, YOLO \cite{terven_comprehensive_2023} divides the input image into a grid and predicts bounding boxes and class probabilities for each grid cell. 
Although this efficiency is beneficial for real-time applications, it often comes at the cost of accuracy when compared to two-stage detectors

\paragraph{Two-stage detectors}These models, among which Faster R-CNN \cite{ren_faster_2017} can be considered to be the most representative one, follow a more complex approach that divides the detection process into two stages. In the first stage, a Region Proposal Network (RPN) generates a set of candidate object proposals (bounding boxes) from the input image.
In the second stage, these proposals are refined and classified into different object categories by a second network. This second stage typically involves a more complex network, such as a convolutional neural network (CNN), which performs classification and further refinement of the bounding box coordinates. 
This two-step process boosts accuracy by allowing for a more refined feature analysis, though it also slows down processing, making two-stage detectors less suited for applications that require high-speed performance.

\vspace{2mm}
Most detector networks, including Faster R-CNN and YOLO, produce a large number of bounding box proposals which are subsequently refined using confidence thresholding and Non-Maximum Suppression (NMS) to produce a set of finally detected objects in the image. 
Each bounding box proposal $d_i$ can be defined as follows:
\begin{equation} \label{eq:object_detectors_output}
\mathbf{d}_i = \left[\mathbf{L}_i, O_i, \mathbf{P}_i\right] = \left[(x^i_{1}, y^i_{1}, x^i_{2}, y^i_{2}), O_i, (p^i_{1}, \ldots, p^i_{C})\right],
\end{equation}
where $\mathbf{L}_i$ defines the bounding box corners \((x^i_{1}, y^i_{1})\) and \((x^i_{2}, y^i_{2})\); $O_i \in [0, 1]$ refers to the probability that bounding box $L_i$ contains an object of any class; and $\mathbf{P}_i$ is a vector of probabilities \((p^i_{1}, \ldots, p^i_{C})\) representing the probability that region \( \mathbf{L}_i \) belongs to each of \( C \) classes. 
Unlike traditional classifiers, which assign a single class label to an entire image, object detectors must handle both classification and localization simultaneously. This dual task, predicting the class and precise location of each object, increases the complexity of making these models interpretable.

\subsection{Explainable Artificial Intelligence (XAI)}\label{backgroud:xai}
Despite the great performance exhibited by object detectors in manifold applications, their adoption in risk-sensitive scenarios is often hindered by a lack of trust and transparency by the user making decisions based on the detections issued by these models. As introduced previously, research on XAI produce techniques and methods that make the behavior and predictions of AI models understandable to humans without sacrificing performance \cite{gunning_xaiexplainable_2019}. To this end, multiple XAI techniques have been proposed, which can be classified into four broad categories \cite{ali_explainable_2023}:
\begin{itemize}[leftmargin=*]
    \item \textit{Scoop-based techniques} focus on the extent of the explanation, providing either local explanations for specific predictions or global explanations for the overall model behavior.
    \item \textit{Complexity-based methods} consider the complexity of the model, with simpler, interpretable models offering intrinsic interpretability and more complex models requiring post-hoc explanations.
    \item \textit{Model-based approaches} distinguish between XAI methods that are specific to particular types of models, and those that are model-agnostic, capable of being applied to any model disregarding the specifics of their internals.
    \item \textit{Methodology-based techniques} are categorized by their methodological approach, such as backpropagation-based methods that trace input influences, or perturbation-based methods that alter inputs to observe changes in the output of the model.
\end{itemize}

Given that object detectors are typically complex neural networks, they fall under the \emph{complexity-based} category, thereby requiring post-hoc explainability methods to explain their decisions. Among the various \emph{methodology-based} techniques, \emph{attribution methods} are commonly used to estimate the relevance of each pixel in an image for the detection task. Attribution methods are particularly important for object detection, where both localization and classification need to be explained.

Traditional attribution methods have been primarily developed for image classifiers \cite{abhishek2022attribution}, which produce a single categorical output, making them less suited for object detectors. Object detectors, unlike classifiers, generate multiple detection vectors that encode not only class probabilities, but also localization information and additional metrics, such as objectness scores (see Section \ref{backgroud:objdet}). Furthermore, techniques like NMS and confidence threshold filtering, which are used to refine bounding box proposals, add complexities that require a deeper understanding of the model’s internal workings, complicating the use of certain XAI methods, such as gradient-based approaches. Therefore, we focus on \emph{model-agnostic black-box XAI} approaches, which are designed to be architecture-independent, and do not depend at all on the specifics of the model under target.

Among model-agnostic XAI methods, \emph{perturbation-based} approaches are commonly used due to their simplicity and effectiveness in revealing which parts of the input are most influential for the model's predictions. Perturbation-based techniques offer a direct way to assess how changes to the input image affect the model's output. By systematically altering or masking parts of the input image (using masks to generate perturbed samples), these methods allow inferring the importance of different regions based on the model’s input-output behavior.

\begin{figure}[h!]
    \centering
    \includegraphics[width=0.9\linewidth]{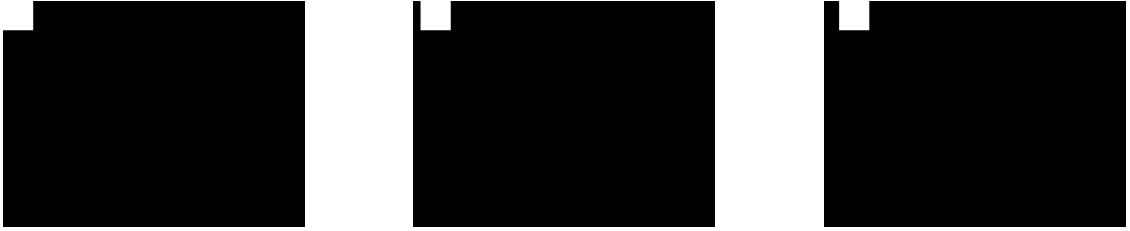}
    \includegraphics[width=0.9\linewidth]{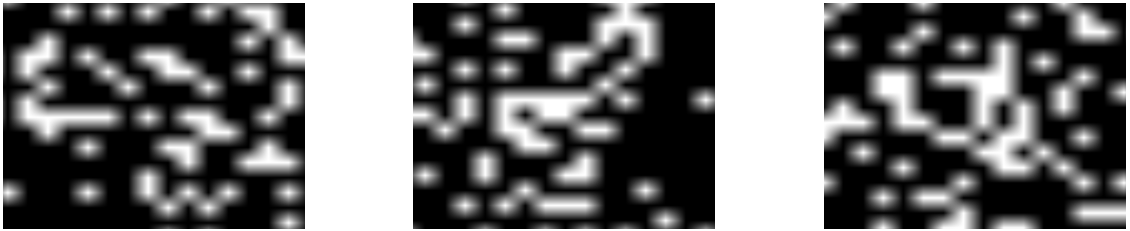}
    \includegraphics[width=0.9\linewidth]{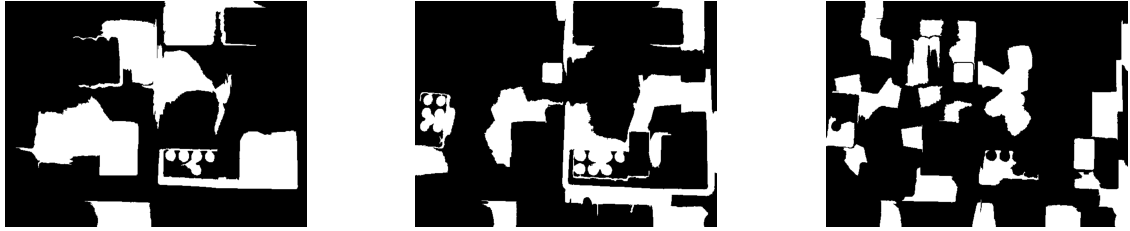}
    \caption{Example of three masks generated using Sliding Window (top), RISE (middle), and MFPP (bottom). MFPP masks are dependent on the image at the input of the model. In this case, we consider a sample from the battery assembly dataset detailed in Section \ref{sec:materials_and_methods}.}
    \label{fig:masks_slidingw_rise_mfpp}
\end{figure}

The typical pipeline for perturbation-based XAI methods can be divided into three stages: \emph{(1) Data Preparation, (2) Model Assessment, and (3) Importance Computation}. In the Data Preparation stage, masks are generated and applied to the image to create perturbed samples. The Model Assessment stage involves passing these perturbed images through the model to observe the changes in output. Finally, in the Importance Computation stage, the importance of each pixel is calculated by comparing the model's outputs for the original and perturbed images. While the Model Assessment stage remains consistent across methods, with each perturbed image passed through the model, the Importance Computation varies depending on the XAI approach used. This can range from simple techniques like retraining a model (e.g., LIME) to more complex approaches. Since the effectiveness of these methods largely depends on how the perturbed images are generated, three mask generation algorithms are next described (Figure \ref{fig:masks_slidingw_rise_mfpp}):
\begin{itemize}[leftmargin=*]
\item \emph{Sliding Window}: This method, which is similar to the Occlusion technique proposed in \cite{fleet_visualizing_2014}, systematically moves a window of fixed size across the image and sets the region within the window to a constant value (e.g., zero) to occlude that part of the image. By iteratively sliding the window across the entire image, we can assess the impact of each occluded region on the model's output. The method requires specifying the window size, which determines the area of the image being occluded at each step, and the stride, which sets how much the window moves between iterations.

\item \emph{RISE}: 
Randomized Input Sampling for Explanation (RISE) \cite{petsiuk_rise_2018} involves sampling \( N \) binary masks of size \( h \times w \), which are smaller than the original image size \( H \times W \). Each element in the mask is independently set to 1 with probability \( p \) and to 0 with the remaining probability \( 1 - p \). These masks are then upsampled to size \( (h + 1)\cdot C_H \times (w + 1)\cdot C_W \) using bilinear interpolation, where \( C_H \times C_W = \left\lfloor H/h \right\rfloor \times \left\lfloor W/w \right\rfloor \). The upsampled masks are cropped to the original image size \( H \times W \) with uniformly random offsets ranging from \( (0, 0) \) to \( (C_H, C_W) \). This method creates a diverse set of masks that cover different parts of the image, allowing for a comprehensive evaluation of the importance of various regions.

\item \emph{MFPP}: The so-called Morphological Fragmental Perturbation Pyramid (MFPP) \cite{yang_mfpp_2020} method divides the input image into multi-scale fragments and perturbs them randomly. In this sense, it is similar to RISE, but instead of perturbing elements of the generated masks with dimension \( h \times w \), MFPP defines regions according to segmentations at different scales. Depending on the number of defined fragments, the regions would be more fine-grained yet more time-consuming. The segments are dependent on each image, requiring the creation of new masks for every image.
\end{itemize}

\section{Materials and Methods}\label{sec:materials_and_methods}

This section describes the industrial robotics use cases in what refers to the datasets (Section \ref{sec:datasets}), object detection model (Section \ref{materials:yolov8}), XAI methods (Section \ref{materials:xai_methods}) and the explanation quality metrics (Section \ref{sec:metrics}) considered in our work. The novel XAI technique and quality metrics proposed in this manuscript are also described in Section \ref{materials:xai_methods}.

\subsection{Industrial Robotics Datasets under Consideration} \label{sec:datasets}
The datasets used in this manuscript have been collected during the course of the ULTIMATE project, \url{https://ultimate-project.eu/}, which features two distinct real robotics use cases \cite{kozik_ultimate_2024}. The first dataset, from PIAP \url{https://piap.lukasiewicz.gov.pl/}), involves a collaborative workspace where a human and a robotic arm work together. The second dataset, provided by Robotnik \url{https://robotnik.eu/}, focuses on a battery assembly area, where a robotic arm assembles components for a battery kit\footnote{While the datasets contain a relatively small number of images, this data shortage is typically encountered in real-world industrial scenarios subject to data availability constraints. Nevertheless, in the use cases under considerations the contextual and scene variability is minimal, yielding short-tailed distributions of the objects to be detected. Therefore, the small datasets described in the paper sufficiently capture the relevant features for the specific object detection tasks addressed by the models.}.

\begin{figure}[h]
    \centering
    \includegraphics[width=0.25\linewidth]{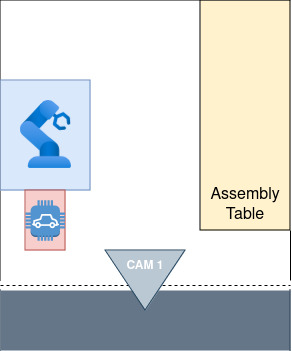}
    \includegraphics[width=0.25\linewidth]{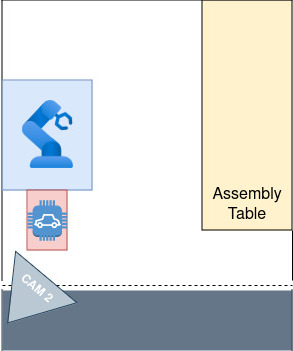}
    \includegraphics[width=0.25\linewidth]{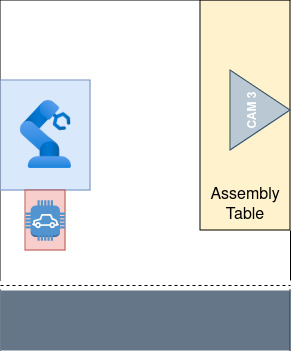}
    \includegraphics[width=0.25\linewidth]{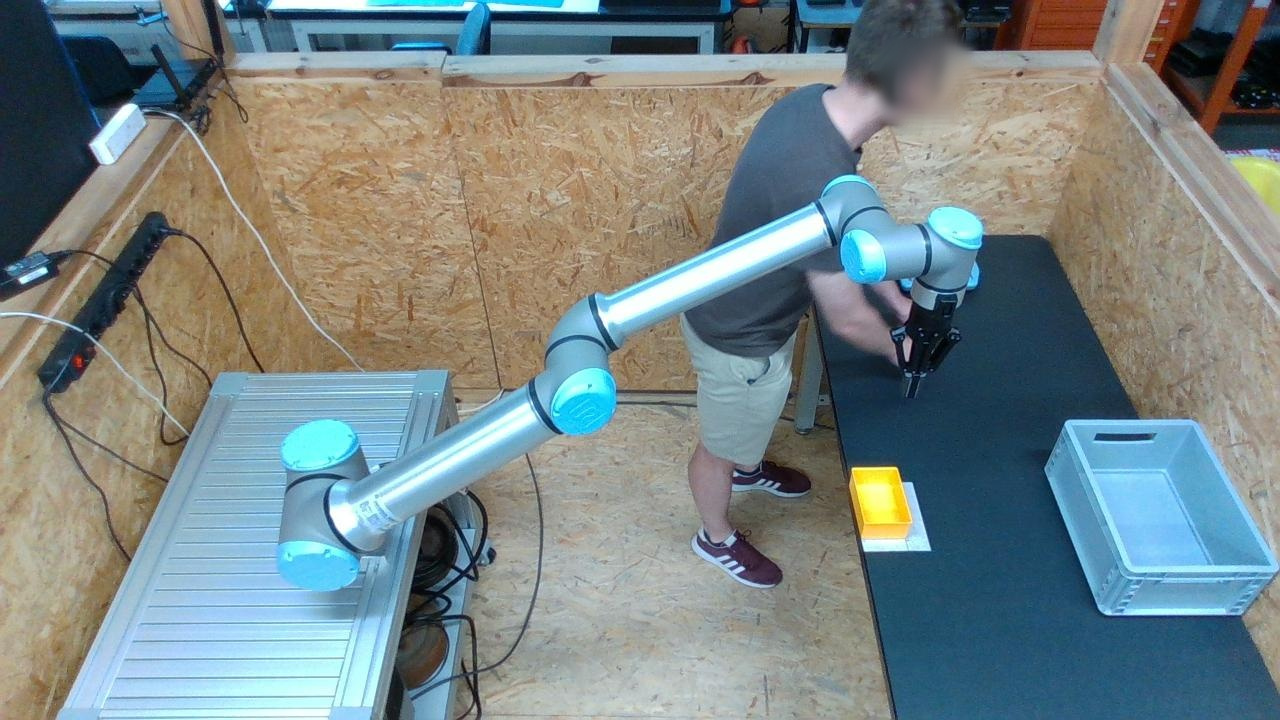}
    \includegraphics[width=0.25\linewidth]{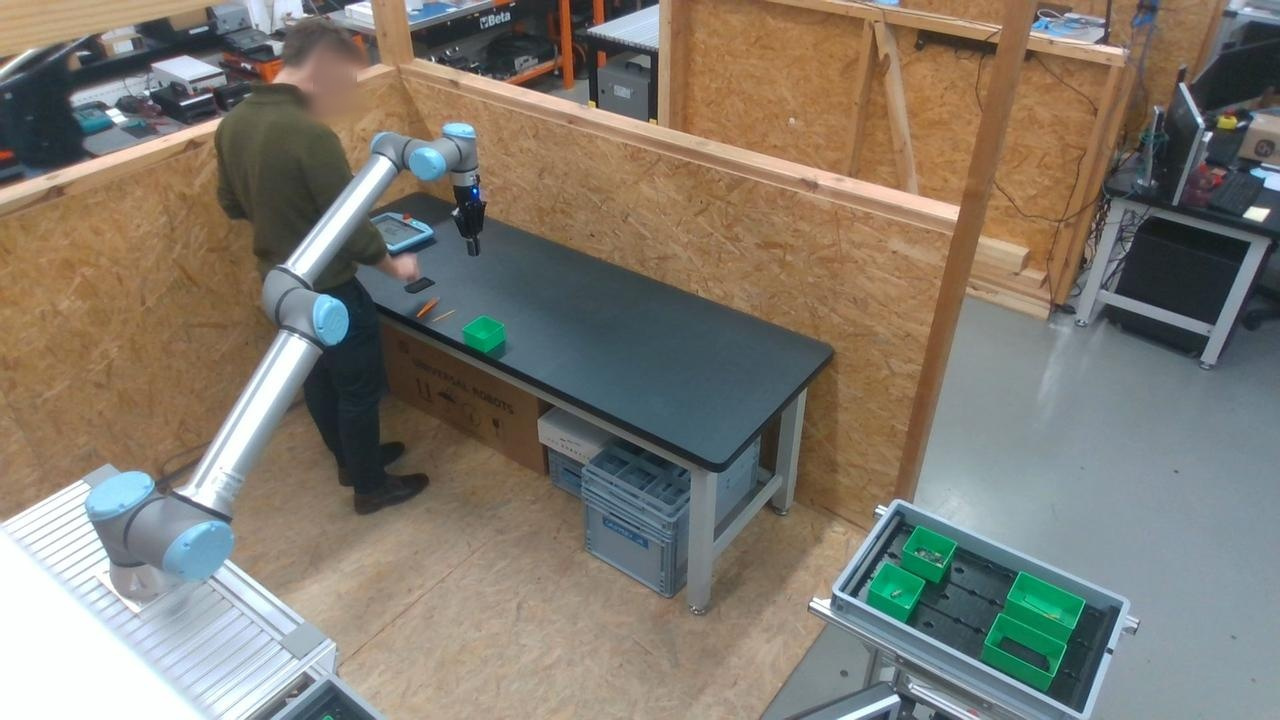}
    \includegraphics[width=0.25\linewidth]{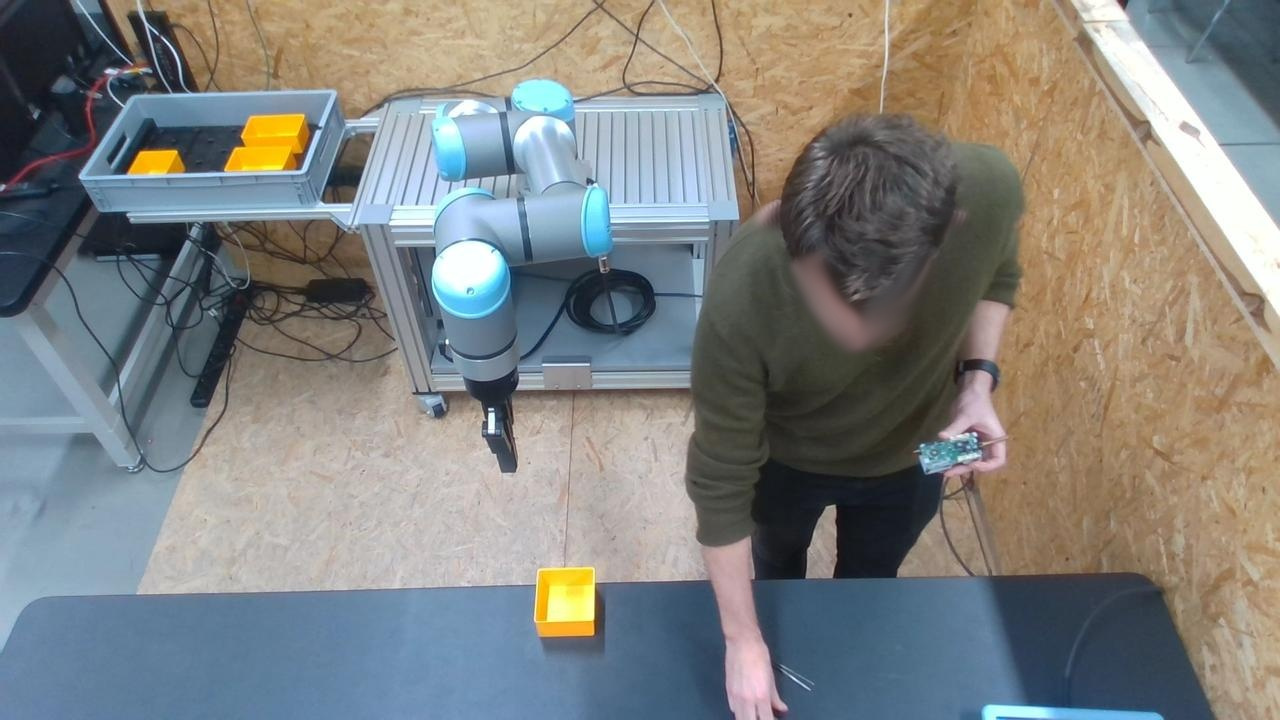}
    \caption{Dataset 1 (Human-Robot collaboration): Data are captured from cameras located in 3 different positions. All the images belonging to this dataset contain the faces blur to preserver anonymity. }
    \label{fig:piap_data_setup}
\end{figure}

\paragraph{Dataset 1: Human-Robot Dataset}This dataset consists of 96 images captured from three different cameras, as exemplified in Figure \ref{fig:piap_data_setup}, with 32 images taken from each camera. The dataset includes two object classes: \texttt{human} and \texttt{gripper}. Importantly, each image in this dataset contains only a single object of each class, meaning a maximum of one human and one \texttt{gripper} per image. To ensure a diverse and representative sample, we applied feature extraction using ResNet \cite{he_deep_2015} to obtain embeddings for the entire dataset. The dimensionality of these embeddings was reduced using Principal Component Analysis (PCA), followed by K-means clustering (with $k=8$ clusters). From each cluster, four images were randomly selected, resulting in a final subset. The data were split into three sets: 72 images for training (75\%), 6 for validation (6.25\%), and 18 for testing (18.75\%). To maintain consistency, we applied the same partitioning to the data from each camera. This resulted in 24 images for training, 2 for validation, and 6 for testing from each camera.

\begin{figure}[h]
    \centering
    \begin{tabular}{c}
        \includegraphics[width=0.455\linewidth]{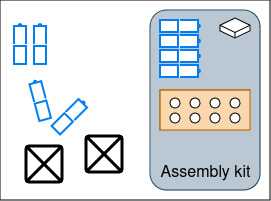}\hspace{3mm}
        \includegraphics[width=0.45\linewidth]{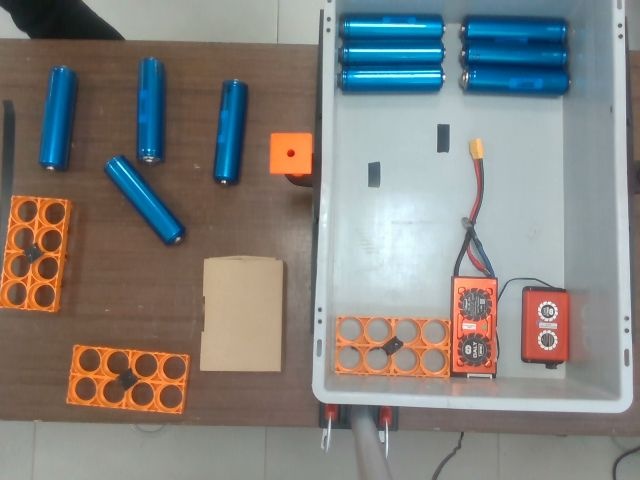}
    \end{tabular}
    \caption{Dataset 2 (Battery Assembly kit): The setup where a robotic arm would assemble the kit based a bird-eye view of the table where all component are expected to be; (left) a theoretical setup; (right) an actual sample.}
    \label{fig:rob_data_setup}
\end{figure}

\paragraph{Dataset 2: Battery Assembly Dataset} This dataset consists of 7 images, all captured from a bird's-eye (top-down) view, showing a robotic arm assembling a battery kit, as shown in Figure \ref{fig:rob_data_setup}. The dataset includes five distinct object types: \texttt{individual battery}, \texttt{bms\_a}, \texttt{bms\_b}, \texttt{battery holder}, and \texttt{unknown object}. In contrast to the Human-Robot Dataset, each image in the Battery Assembly Dataset may contain multiple objects of the same class, such as several individual batteries in a single scene.

\vspace{2mm}
It is worth noting that XAI techniques can be applied to any type of data. When applied to training data, they help reveal what the model has learned to focus on during training. When applied to test data, they provide insight into how well the model generalizes to new, unseen examples. For the Human-Robot Dataset, XAI explanations were applied exclusively to the test images, allowing us to assess the model's behavior on unseen data. However, for the Battery Assembly Dataset, given the limited number of images (only 7), XAI explanations were applied to the entire dataset.

\subsection{Object Detection Model: YOLOv8}\label{materials:yolov8}
Among the possible object detector models, we selected one of the state-of-the-art options, YOLOv8, due to its numerous advancements over previous versions and its robust performance in object detection tasks \cite{terven_comprehensive_2023}. 
YOLOv8 \cite{reis_real-time_2024} integrates a novel combination of Feature Pyramid Network (FPN) and Path Aggregation Network (PAN) architectures, enhancing its ability to detect objects at various scales and resolutions. The FPN gradually reduces the spatial resolution of the input image while increasing feature channels, facilitating multi-scale object detection. The PAN architecture further aggregates features from different levels through skip connections, improving the detection of objects with diverse sizes and shapes. Additionally, YOLOv8 introduces an anchor-free detection mechanism that directly predicts the center of an object (instead of the offset from a known anchor box), reducing the number of box proposals and speeding-up the post-processing. Furthermore, it was trained with larger and more diverse datasets including the popular COCO dataset, improving its performance across a wider range of images.

YOLOv8 was developed and released by Ultralytics, and although the model and its weights are open-source, most users are expected to utilize the Ultralytics framework for its enhanced usability. However, unlike previous YOLO releases where the probability for each class per predicted box was accessible, in YOLOv8, the Ultralytics API outputs only the probability for the class with the highest confidence in each box\footnote{\url{https://github.com/ultralytics/ultralytics/issues/2863}\\ \url{https://github.com/ultralytics/ultralytics/issues/4908}}. Consequently, by default, YOLOv8 outputs:
\begin{equation}
\mathbf{d}_i = \left[\mathbf{L}_i, O_i, C_i\right] = \left[(x^i_{1}, y^i_{1}, x^i_{2}, y^i_{2}), O_i, C_i\right],
\end{equation}
where $\mathbf{L}_i=(x_1^i,y_1^i,x_2^i,y_2^i)$ represents the coordinates of the bounding box, $O_i$ denotes the objectness score, and $C_i$ corresponds to the predicted class label for the object within the bounding box, which differs with respect to the outputs shown in Expression \eqref{eq:object_detectors_output}.

\subsection{Explainability Methods} \label{materials:xai_methods}

We evaluate four popular methods for generating visual explanations of black-box models: LIME, RISE, D-RISE, and D-MFPP. The first two methods, LIME and RISE\footnote{These XAI methods have been chosen due to their perturbation-based nature, which aligns closely with the methodology followed by the XAI methods D-RISE and D-MFPP proposed in this work. Both D-RISE and D-MFPP generate explanations through perturbations.}, were originally developed for image classifiers but can be adapted to object detectors, However, they primarily focus on explaining classification aspects and are not capable of addressing localization characteristics. In contrast, D-RISE is one of the first XAI methods specifically designed for object detectors, providing explanations that encompass both classification and localization. Additionally, we extend the existing MFPP method (originally tailored for classifiers) into a version suitable for object detection, which we refer to as D-MFPP. In what follows we briefly describe them, flowing into a description of the proposed D-MFPP approach:
\begin{itemize}[leftmargin=*]
\item \textbf{LIME} was originally designed to explain the predictions of any classifier by approximating it locally with an interpretable model. To explain the prediction for an input image $I$, LIME fits an interpretable model $g$ (e.g., a linear model) to approximate the behavior of the black-box model $f$ locally around $I$. The similarity between the original image and the perturbed samples is measured using a kernel function $\pi_I(z)$. When image explanations are targeted, LIME groups contiguous pixels into superpixels based on similar features they represent. This approach allows LIME to measure the importance of regions in the image rather than individual pixels, making the explanations more interpretable. 

\item As introduced in the previous section, \textbf{RISE} \cite{petsiuk_rise_2018} was originally designed for deep neural networks that take images as input and output a class probability (e.g., a classifier like ResNet-50). It generates saliency maps that indicate the importance of each pixel by applying randomly generated binary masks \(M_i\) to the input image \(I\) and observing the changes in the model's output \(f(I \odot M_i)\). In RISE, \(N\) binary masks \(M_i \in \{0,1\}^{h \times w}\) are generated (as explained in Section \ref{backgroud:xai}). 
These masks are then applied to the input image \(I\) to generate masked images \(I_i' = I \odot M_i\), where \(\odot\) denotes element-wise multiplication. The model is evaluated on each masked image \(I_i'\) to obtain the outputs \(f(I \odot M_i)\). The importance score for each pixel \((x,y)\) is then calculated as the weighted sum of the outputs:
\begin{equation} \label{eq:rise}
    S_{I,f}(x,y) = \frac{1}{N} \sum_{i=1}^{N} f(I \odot M_i) \cdot M_i(x,y)    
\end{equation}
where the weights \(M_i(x,y)\) represent the value of mask \(i\) at pixel \((x,y)\). The intuition behind RISE is that $f(I \odot M_i)$ would be high when pixels preserved by mask $M_i$ are important. Although this is true when having infinite diverse masks, in practice RISE calculates each pixel's importance empirically by Monte Carlo sampling. Therefore, RISE largely depends on the number of masks ($N$) and how they are generated (i.e., is sensitive to  the selected probability $p$ and resolution $s$).
\end{itemize}

\subsubsection{D-RISE and Proposed D-MFPP Approach}
Unlike the other two approaches originally designed for classifiers that measure solely classification aspects, \textbf{D-RISE} (Detector Randomized Input Sampling for Explanation) \cite{petsiuk_black-box_2021} was designed to explain both the classification and localization of a detection. 
In this sense, D-RISE extends RISE by producing saliency maps specifically for object detectors. As previously seen in Section \ref{backgroud:objdet}, the output given by an object detector differs from the probability vector given by a classifier, obtaining localization information $L_i$, an objectness score $O_i$ and the probability of classifying each bounding box to any of the considered classes $P_i$. As a consequence, Expression \eqref{eq:rise} used by RISE is replaced in D-RISE with a new similarity score, given by:
\begin{equation}
    S_{I,f}(\mathbf{d}_t,\mathbf{d}_j) = s_L(\mathbf{d}_t,\mathbf{d}_j) \cdot s_P(\mathbf{d}_t,\mathbf{d}_j) \cdot s_O(\mathbf{d}_t,\mathbf{d}_j),
\end{equation}
where $s_L = IoU(\mathbf{L}_t, \mathbf{L}_j)$, $s_P = \mathbf{P}_t \cdot \mathbf{P}_j/(||\mathbf{P}_t||\cdot ||\mathbf{P}_j||)$, and $s_O = O_j$. In this formulation, $s_L$ represents the spatial proximity of the bounding boxes encoded by the target detection $\mathbf{d}_t$ and the proposal $\mathbf{d}_j$, measured using the Intersection over Union (IoU); the term $s_P$ evaluates the similarity between the class probabilities of the target detection and the proposal using cosine similarity; and $s_O$ incorporates the objectness score of the proposal $O_j$. 
It is important to note that for a detection target $\mathbf{d}_t$ there would potentially be more than one detection proposals $\mathbf{d}_j$.
Therefore, we would have multiple $S_{I,f}(\mathbf{d}_t,\mathbf{d}_j)$. As explained in D-RISE, the explanations consider only the detection with maximal score for each mask:
\begin{equation}
    S_{I,f}\left(\mathbf{d}_t, f(M_i \odot I)\right) = \max_{\mathbf{d}_j \in f(M_i \odot I)} S_{I,f}(\mathbf{d}_t, \mathbf{d}_j).
\end{equation}

Given the YOLOv8 outputs explained in Section \ref{materials:yolov8}, which do not provide the class probability vector $P_i$ without modifying its architecture (an approach we want to avoid within the scope of this paper), we must adapt the similarity score to only consider $s_L$ and $s_O$. Consequently, the modified similarity score can be expressed as:
\begin{equation}\label{eq:drise_adapted}
S_{I,f}(\mathbf{d}_t,\mathbf{d}_j) = s_L(\mathbf{d}_t,\mathbf{d}_j) \cdot s_O(\mathbf{d}_t,\mathbf{d}_j) = IoU(\mathbf{L}_t, \mathbf{L}_j) \cdot O_j.
\end{equation}

This adjustment allows still utilizing D-RISE effectively for generating saliency maps with the default YOLOv8 model, focusing on the spatial and objectness aspects of detections, while maintaining the integrity of the model's original architecture.

Similarly, we can adopt this similarity score but apply it with a different mask generation process. The MFPP method introduced in Section \ref{backgroud:xai}, originally designed for classification tasks, can be extended by applying Equation \eqref{eq:drise_adapted}, resulting in \emph{D-MFPP}. To the best of our knowledge, no previous work has proposed this variant of MFPP for object detection tasks.

\subsection{Metrics} \label{sec:metrics}
Evaluating the performance of attribution-based explainability methods for image data involves assessing how well the generated relevance heatmaps highlight important regions of the input image that contribute to the model's decision. Generally, according to \cite{hedstrom_quantus_2023}, explanation quality metrics can be grouped into six categories based on their logical similarity: faithfulness, robustness, localization, complexity, randomization, and axiomatic metrics. In this study, we focus on two of these categories that are particularly relevant to object detection: localization (Section \ref{sssec:local}) and faithfulness (Section \ref{sssec:faith}).

\subsubsection{Localization} \label{sssec:local}

Localization metrics evaluate whether the explainable evidence is centered around a region of interest (RoI) defined by a bounding box, segmentation mask, or a cell within a grid. These metrics aim to verify if the saliency maps correctly highlight the areas in the image that contain the object of interest. Among them, our experiments will consider:
\begin{itemize}[leftmargin=*]
\item \emph{Pointing Game} (PG), which is a human evaluation metric introduced in 
\cite{zhang_top-down_2018}. If the highest saliency point lies inside the human-annotated bounding box of an object, it is counted as a hit. The PG accuracy is given by:
\begin{equation}
    \text{PG} = \frac{\# Hits}{\# Hits + \# Misses},
\end{equation}
which is averaged over all categories in the dataset.

\item \emph{Energy-based Pointing Game} (EBPG) \cite{wang_score-cam_2020}, which measures the proportion of activations within the given bounding box relative to the whole activation in the image. It assesses how much of the model's activation energy is concentrated within the predefined region of interest. Formally:
\begin{equation}
\text{EBPG} = \frac{\sum_{(x,y) \in \text{bbox}} S_{I,f}(x,y)}{\sum_{(x,y) \in \text{bbox}} S_{I,f}(x,y) + \sum_{(x,y) \notin \text{bbox}} S_{I,f}(x,y)},
\end{equation}
where \smash{\( S_{I,f}(x,y) \)} represent the saliency score at pixel \smash{\((x,y)\)}, \smash{\( \sum_{(x,y) \in \text{bbox}} S_{I,f}(x,y) \)} represents the sum of activation values within the bounding box, and \( \sum_{(x,y) \notin \text{bbox}} S_{I,f}(x,y) \) represents the sum of activation values outside the bounding box.
\end{itemize}

\subsubsection{Faithfulness} \label{sssec:faith}
\begin{figure}[!ht]
    \centering
    \includegraphics[width=0.9\linewidth]{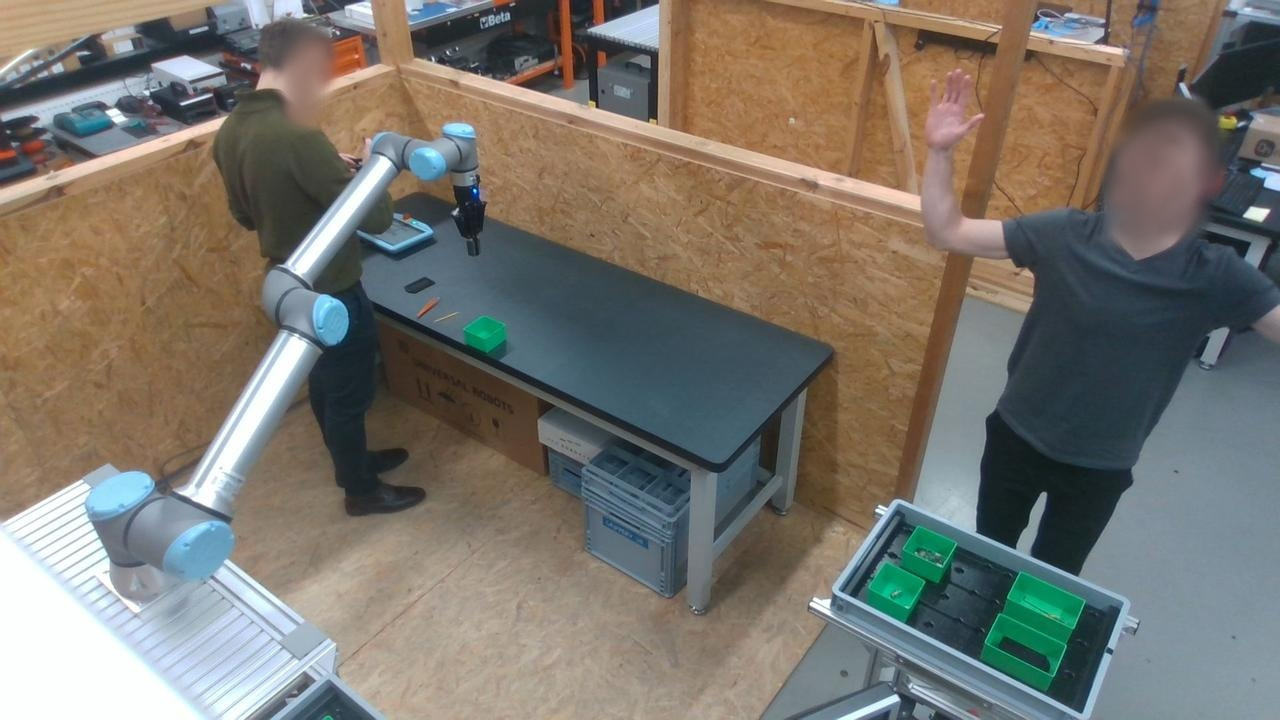}
    \\
    \includegraphics[width=0.45\linewidth]{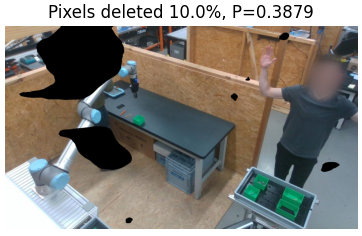}
    \includegraphics[width=0.45\linewidth]{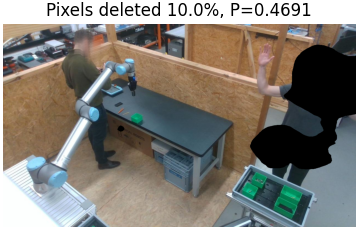}
    \\
    \includegraphics[width=0.45\linewidth]{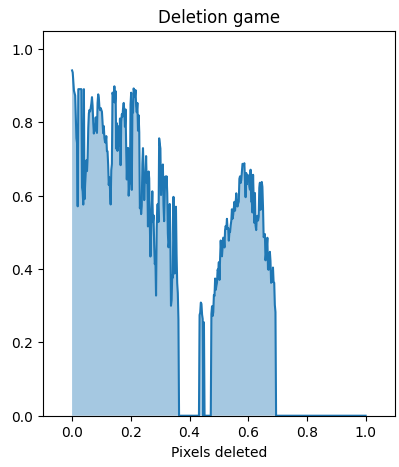}
    \includegraphics[width=0.45\linewidth]{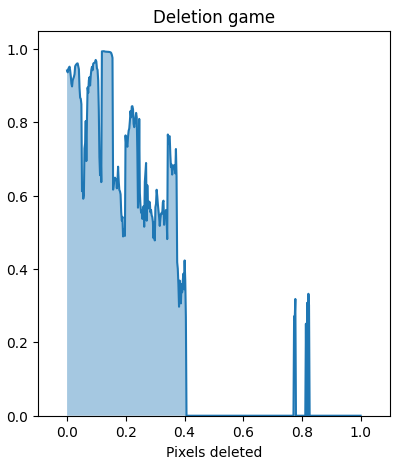}
    \\
    \includegraphics[width=0.45\linewidth]{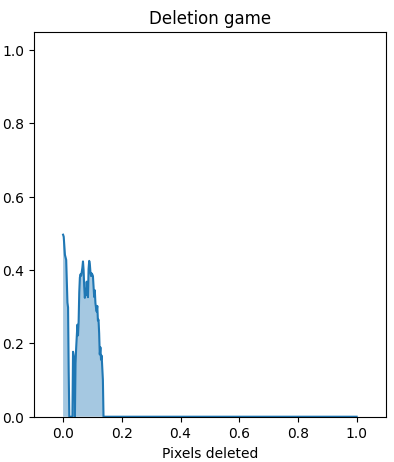}
    \includegraphics[width=0.45\linewidth]{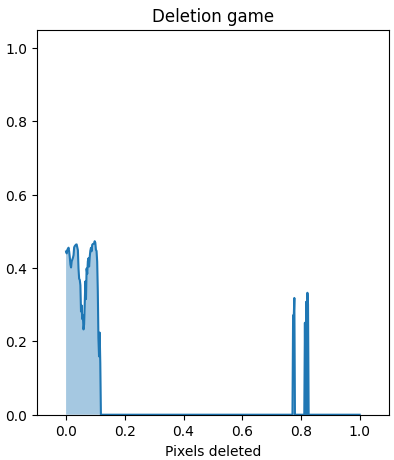}
    \\
    \caption{Illustration of a collaborative workspace featuring two humans and a robotic arm. The first row shows the original image. The second row displays the image with the 10\% most important pixels removed for each human, as identified by an XAI method. In the third row, the Deletion metric curve, which only considers class type, shows a high probability score even when the primary human is largely occluded by the other person. The fourth row presents the D-Deletion metric curve, which incorporates a localization component, providing a more accurate measure of explanation importance by considering the positions of entities within the image. A lower area under the curve indicates a better explanation.}
    \label{fig:d-deletion_motivation}
\end{figure}
Metrics accounting for faithfulness quantify to what extent explanations follow the predictive behavior of the model, asserting that more important features play a larger role in model outcomes. These metrics focus on understanding the causal relationship between input features and the model's output by systematically altering the features and observing the changes in predictions. Among them:
\begin{itemize}[leftmargin=*]
\item \emph{Deletion}: Inspired by the work by 
\cite{bach_pixel-wise_2015}, the Deletion metric was proposed in RISE \cite{petsiuk_rise_2018}. This metric measures a decrease in the probability of the predicted class as more and more important pixels are removed, where the importance is obtained from the saliency map. A sharp drop, and thus a low Area Under the probability Curve (AUC, as a function of the fraction of removed pixels), indicates a good explanation. Given the importance score for each pixel calculated by any XAI method, $S_{I,f}$, we can formulate the Deletion metric as:
\begin{equation}\label{eq:original_del}
    \text{Deletion(I,S,c)} = \textup{AUC} \left( \left\{Pr(f \left( I \odot M_k \right)=c) \right\}_{k=1}^{K} \right),     
\end{equation}
where $I$ is the original image, $M_k$ represent a  mask with the the $k$-th most important pixels removed sorted by $S_{I,f}$, $Pr(f \left( I \odot M_k \right)=c)$ represents the probability of model $f$ predicting that the bounding box belongs to class $c$, and AUC($\cdot$) computes the area under the curve for the $K$ predictions.

\item \emph{Minimum Subset}: It follows the same logic as \textit{Deletion}, but instead of determining the AUC, it considers the required number of pixels that make the prediction to change \cite{gevaert_evaluating_2024}. Given the importance score for each pixel ($S_{I,f}$), \textit{Min-Subset} is defined as the smallest subset of pixels that needs to be removed to change the model's prediction. Mathematically:
\begin{align}
&\textup{Min-Subset}(I, S, c) = \nonumber \\
&\min \left\lbrace k\in\{1,2\ldots,K\}:\: f \left(I \odot M_k \right) \neq f \left (I\right)\right\rbrace,
\end{align}
where $f(I \odot M_k)$ represents the class label assigned by the model $f$ after passing the image $I$ with the top $k$ most important pixels removed, and $f(I)$ is the class label predicted for the original image.

\end{itemize}

\subsubsection{Proposed D-Deletion and D-Minimal Subset Metrics}

\begin{algorithm}
\caption{Deletion Metric's Pseudocode for Object Detector}
\label{alg:pseudocode_deletion}
\begin{algorithmic}[1]
\REQUIRE Image $I$, saliency map $S_{I,f}$, number of steps $K$, target detection $\mathbf{d}_t$, target class $C_t$
\STATE Initialize $S \leftarrow []$

\FOR{$k = 1$ to $K$}  
    \STATE $M_k \leftarrow S_{I,f}$ removing the top $k$ most important pixels
    \STATE Apply mask $M_k$ to image $I$ 
    \STATE Forward pass through the model $f$ and obtain the bounding box proposals $\mathbf{d}_j = [\mathbf{L}_j, O_j, C_j] = f(I \odot M_k)$
    \STATE Initialize list of proposals: \texttt{proposals} $\leftarrow []$
    
    \FOR{each bounding box $\mathbf{d}_j$ predicted by the model $f$}
        \IF{$C_j = C_t$}
            \STATE \texttt{proposals.append}($O_j$)
            \begin{tcolorbox}[colback=lightgray!30, boxrule=0pt, sharp corners, sharp corners, colframe=white, left=0pt, right=0pt, top=0pt, bottom=0pt, boxsep=0pt, width=\linewidth]
            \% \textit{For D-Deletion}
            \IF{ $IoU(\mathbf{d}_t,\mathbf{d}_j)> \gamma$ }
                    \STATE \texttt{proposals.append}($O_j$)
                \ENDIF
            \end{tcolorbox}
        \ELSE
            \STATE \texttt{proposals.append}(0)
        \ENDIF
    \ENDFOR
    
    \STATE Insert the maximum score within the deletion buffer: $S \leftarrow S \cup \max(\texttt{proposals})$
\ENDFOR

\STATE Calculate the Deletion metric as the area under the curve: $D = \text{AUC}(S)$
\RETURN Deletion score $D$
\end{algorithmic}
\end{algorithm}

Originally, Deletion was designed for classifiers. However, with object detectors, multiple detections in a single image can occur. Although D-RISE stated the necessity to adapt this metric for object detectors \cite{petsiuk_black-box_2021}, no formal definition can be found in the literature. 
Therefore, considering the importance of this issue in real use cases, we formally re-define Equation \eqref{eq:original_del} in two manners:
\begin{enumerate}[leftmargin=*]
    \item \textit{Deletion}. Measures the explanation given the target class label $C_t$ (regardless if there is more than one element for a class) and iteratively removes the top $k$ most important pixels:
    \begin{align}\label{eq:deletion_allclases}
    &\text{Deletion}(I, S, C_t)= \nonumber \\
    & AUC \left( \left\{ \max_{\mathbf{d}_j^k} \left[ O_j^k \cdot \mathds{I}\{C_j^k = C_t\} \right] \right \}^K_{k=1} \right),
    \end{align}

    The model \( f (\cdot)\) takes as input the masked image \( I \odot M_k \) and outputs a set of bounding box proposals \( \mathbf{d}_j^k = [\mathbf{L}_j^k, O_j^k, C_j^k] \). The indicator function \smash{\( \mathds{I}\{C_j^k = C_t\} \)} equals 1 if the predicted class \( C_j^k \) matches the target class \( C_t \), and 0 otherwise. The term \smash{\( \max_{\mathbf{d}_j^k} \left[ O_j^k \cdot \mathds{I}\{C_j^k = C_t\} \right] \)} selects the maximum objectness score \( O_j^k \) for the bounding boxes where the predicted class matches the target class. The AUC is then computed over the set of prediction scores for the $K$ steps, where at each step the most important pixels are progressively removed.
    
    \vspace{1mm}
    \item \textit{D-Deletion}. 
    While the standard Deletion metric evaluates the impact of pixel removal on a class prediction, it lacks the ability to account for spatial localization, which is essential in object detection tasks where multiple instances of the same class can appear. D-Deletion addresses this limitation by focusing on a specific target bounding box $\mathbf{d}_t$, considering both the class information, $C_t$, and IoU between the target and other detected proposals, $\mathbf{d}_j^k$. This ensures that the metric not only measures faithfulness but also takes localization into account, providing more precise explanations in situations where different objects of the same class coexist. Mathematically is expressed as:
    \begin{align} \label{eq:d-deletion}
        &\hspace{-4mm}\text{D-Deletion}(I, S, C_t)= \nonumber \\ 
        &\hspace{-4mm}AUC \left( \left\{ \max_{\mathbf{d}_j^k} \left[ O_j^k \cdot \mathds{I}\{C_j^k = C_t\} \cdot \mathds{I}\{IoU(\mathbf{d}_t,\mathbf{d}_j^k)>\gamma\} \right] \right \}^K_{k=1} \right)
    \end{align}
    where $\gamma$ is a threshold. As a consequence, when multiple elements of the same class are in an image, \textit{D-Deletion} will only consider those proposals $\mathbf{d}_j^k$ predicted by the model that have a predefined IoU with the target bounding box $\mathbf{d}_t$.
\end{enumerate}

\vspace{1mm} 
The difference between Deletion and D-Deletion is illustrated in Figure \ref{fig:d-deletion_motivation}. This figure highlights how D-Deletion distinguishes between different objects of the same class by incorporating localization information, leading to more refined and accurate explanations ($\downarrow$ AUC in the Figure's last row) when multiple objects of the same class are detected in an image. For the sake of clarity, we provide the pseudocode of Deletion in Algorithm \ref{alg:pseudocode_deletion}, where the main difference with respect to D-Deletion are lines \textbf{10} to \textbf{12}.

\vspace{2mm}
Lastly, akin to how \textit{Min-Subset} is related to \textit{Deletion}, \textit{D-Min-Subset} is  associated with \textit{D-Deletion}. Consequently, \textit{D-Min-Subset} considers both the class type and the IoU to determine the number of pixels required to make the prediction to change:
\begin{align}
&\text{D-Min-Subset}(I, S, C_t) = \nonumber \\
&\min \left\lbrace k\in\{1,2\ldots,K\}:\: C_j^k \neq C_t \text{ or } IoU(\mathbf{d}_t,\mathbf{d}_j^k) < \gamma \right\rbrace,
\end{align}
where $C_j^k$ represents the predicted class label for detection $j$ when passing the masked image $I \odot M_k$ through the model $f$, with $\mathbf{d}_j^k = f(I \odot M_k)$ being the set of detections after removing the top $k$ most important pixels. In this context, \textit{D-Min-Subset} depends on two conditions: (1) the class probability labels $C_j^k$ for the predicted bounding box $\mathbf{d}_j^k$ must no longer match the target class $C_t$, or (2) the IoU between the target bounding box $\mathbf{d}_t$ and the predicted bounding box $\mathbf{d}_j^k$ falls below the threshold $\gamma$. The minimum $k$ is identified as the step where either of these conditions is first met.

\section{Experiments and Results}\label{sec:results}

Contrarily to most studies in the XAI literature that primarily focus on benchmark datasets, our research work focuses on assessing the explainability of object detectors in real-world industrial data. In this context, to evaluate the effectiveness of explanations, we formulate four key research questions to answer them with empirical evidence:
\begin{itemize}[leftmargin=*]
    \item \textbf{RQ1:} Which XAI method provides the most reliable and insightful explanations for object detection models?

    \item \textbf{RQ2:} Does the D-Deletion metric enhance the trustworthiness of XAI outputs when multiple objects of the same class are present in the image?

    \item \textbf{RQ3:} How does the mask generation process influence the quality of explanations, particularly when using similarity scores for object detection? How does D-MFPP behave?

    \item \textbf{RQ4:} Do different image dimensions impact the explanations generated by XAI methods? Do models of varying sizes (large, medium, small, nano) focus on different regions of the image in their explanations?
\end{itemize} 

Next, we outline the hyperparameters used across our experiments to ensure consistency in training and evaluation. For both datasets, models were trained using the YOLOv8 architecture for a total of 100 epochs. The image size (imgsize) was set to the largest dimension of the input image (e.g., $720\times1280 \xrightarrow{} 1280$), and data augmentation techniques such as random horizontal flipping and color jitter were applied. For consistency, the default Ultralytics settings were used wherever applicable. In the case of LIME, we use the baseline implementation of \cite{ribeiro_why_2016}, where we adopt the SLIC segmentation algorithm \cite{achanta_slic_2010} (with 100 segments) and generated 1000 samples to assess the quality of the produced explanations. For RISE and D-RISE, we employed 5000 masks with a probability of 0.25 and a resolution of $16 \times 16$ to produce the saliency maps. Lastly, for all object detection predictions, a confidence threshold of 0.7 was set to determine the validity of each detection.

In what follows we present and discuss on the results obtained to answer each of the RQ formulated above:

\subsection*{RQ1: Comparison between XAI methods}
\begin{figure}
    \centering
    \includegraphics[width=0.45\linewidth]{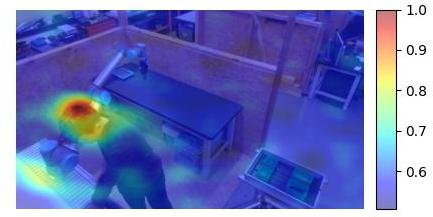}
    \includegraphics[width=0.45\linewidth]{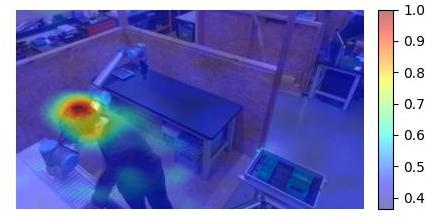}
    \caption{Heatmaps obtained by applying RISE (left) and D-RISE (right) for the detection of a human in the Human-Robot Dataset.}
    \label{fig:rise_vs_drise_piap}
\end{figure}

\begin{figure}[b]
    \centering
    \includegraphics[width=0.45\linewidth]{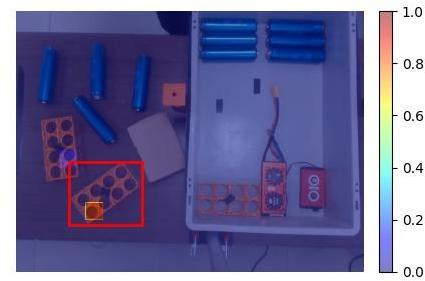}
    \includegraphics[width=0.45\linewidth]{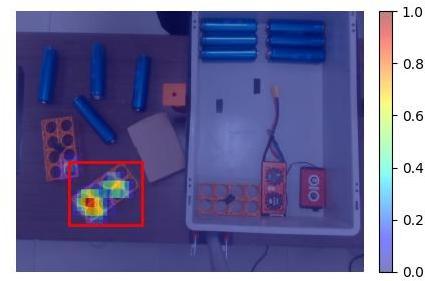}
    \\
    \includegraphics[width=0.45\linewidth]{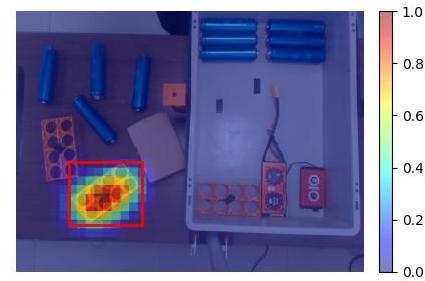}
    \includegraphics[width=0.45\linewidth]{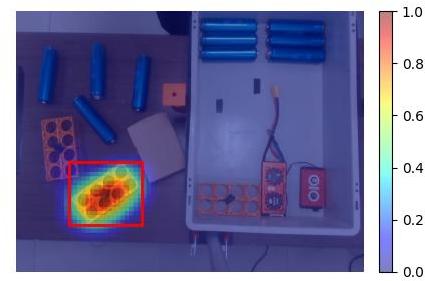}
    \caption{Explanations of a scene using different stride and window size configurations when using D-Sliding Window (combination of mask generation explained in Section \ref{backgroud:xai} and Equation \eqref{eq:drise_adapted}). (Left) Stride of 16; (Right) Stride of 8; (Top) Window size of 32; (Bottom) Window size of 64.}
    \label{fig:slidingwindow_size_stride}
\end{figure}

\begin{table*}[t]
    \centering
    \caption{Quantitative metrics of LIME, RISE and D-RISE for the Human-Robot dataset. The table presents the performance of each XAI technique in terms of classification (Deletion, D-Deletion, Min-Subset, D-Min-Subset) and localization metrics (PG and EBPG), with scores reported for each class (\texttt{Human}, \texttt{Gripper}) and the overall average. 
    Lower values are better for metrics marked with $\downarrow$, while higher values are better for those marked with $\uparrow$. \textbf{Bold values} indicate the best average scores across all objects, highlighting the best-performing XAI method for each metric. Values highlighted in gray represent the best scores for each object category (\texttt{Human} or \texttt{Gripper}) and should be interpreted vertically.
    \vspace{2mm}}

        \begin{tabular}{ccccccccc}
        \toprule
        \textbf{XAI Method} & \textbf{\texttt{Object}} & \textbf{Deletion} ($\downarrow$) & \textbf{D-Deletion} ($\downarrow$) & \textbf{Min-Subset} ($\downarrow$)& \textbf{D-Min-Subset} ($\downarrow$)& \textbf{PG} ($\uparrow$) & \textbf{EBPG} ($\uparrow$) \\
        \midrule
        
        \multirow{3}{*}{LIME} &
        \texttt{Human} & \cellcolor[gray]{0.9} 0.0759 & \cellcolor[gray]{0.9} 0.0632 & \cellcolor[gray]{0.9}4.2703 & \cellcolor[gray]{0.9}4.2703 & \cellcolor[gray]{0.9}1.0000 & \cellcolor[gray]{0.9}34.984 \\
        & \texttt{Gripper} & 0.4688 & 0.0324 & 1.3859 & 1.3859 & \cellcolor[gray]{0.9}1.0000 & \cellcolor[gray]{0.9}2.2270 \\
        \cmidrule(l){2-8}
        & Average & 0.2723 & 0.0478 & \textbf{2.8281} & \textbf{2.8281} & \textbf{1.0000} & \textbf{18.6060} \\
        \midrule
        
        \multirow{3}{*}{RISE} 
        & \texttt{Human} & 0.1827 & 0.1241 & 9.0108 & 9.0108 & 0.7500 & 19.0824 \\
        & \texttt{Gripper} & \cellcolor[gray]{0.9} 0.2637 & 0.0060 & 0.6355 & 0.63 & \cellcolor[gray]{0.9}1.0000 & 1.0542 \\
        \cmidrule(l){2-8}
        & Average & 0.2232 & 0.0651 & 4.8232 & 4.8232 & 0.875 & 10.0683 \\
        \midrule
        
        \multirow{3}{*}{D-RISE} & \texttt{Human} & 0.1255 & 0.0818 & 5.7335 & 5.7335 & 0.8750 & 20.4061 \\
        & \texttt{Gripper} & 0.2777 & \cellcolor[gray]{0.9}0.0059 & \cellcolor[gray]{0.9}0.6091 & \cellcolor[gray]{0.9}0.6091 & \cellcolor[gray]{0.9}1.0000 & 1.0815 \\
        \cmidrule(l){2-8}
        & Average & \textbf{0.2016} & \textbf{0.0438} & 3.1713 & 3.1713 & 0.9375 & 10.7438 \\
        \bottomrule
        \end{tabular}
    \label{tab:piap_rq1}
\end{table*}

In the Human-Robot dataset, the comparison between LIME, RISE, and D-RISE, as shown in Table \ref{tab:piap_rq1}, reveals distinct strengths across different metrics (Section \ref{sec:metrics}).
LIME performs good in terms of localization, with higher PG and EBPG scores (100\% and 18.60\%, respectively) compared to D-RISE (93.75\% and 10.74\%). This indicates that LIME generates more localized saliency maps, focusing closely on the bounding boxes of detected objects. 
However, this superior performance is partly due to the size of the object being analyzed. LIME’s superpixel generation is better suited for larger objects (e.g., \texttt{human}), as larger regions of the image can be grouped effectively into meaningful segments, leading to higher localization scores. This advantage also applies to classification, where larger objects allow LIME to better preserve relevant features for detection. Conversely, for smaller objects (e.g., \texttt{gripper}), LIME struggles when compared to the other methods, as reflected by its worse performance metrics in those cases.

In contrast, RISE and D-RISE are less sensitive to object size, making them more robust across different object scales, which is evident in their better performance on smaller objects like the \texttt{gripper}. They achieve Deletion scores of $0.2636$ and $0.2777$, respectively, compared to LIME's $0.4688$. When considering the overall performance across classes, RISE, with a Deletion score of $0.2232$ and D-Deletion of $0.0651$, shows improvement over LIME in classification-related tasks but still lags behind D-RISE, which achieves the lowest Deletion ($0.2016$) and D-Deletion ($0.0438$) scores. Although D-RISE offers the best balance between classification and localization, the difference between RISE and D-RISE is minimal in this dataset, where each image contains only a single object per class.
As a result, as shown in Figure \ref{fig:rise_vs_drise_piap}, their heatmaps are very similar to each other, both highlighting the human head. However, D-RISE eliminates less relevant areas more effectively.

In the results obtained over the Battery Assembly dataset (Table \ref{tab:rob_rq1}), a similar pattern can be noticed.
LIME excels at localization with an average EBPG of 16.03\%, while RISE and D-RISE perform better in retaining key classification features. Since this dataset includes multiple objects of the same class (e.g., multiple batteries), both LIME and RISE, which are not designed to handle multiple detections of the same class, expose severe limitations. RISE, with a D-Deletion score of $0.1474$, preserves key features better than LIME, but is outperformed by D-RISE, which achieves a score of $0.0344$. D-RISE also shows the highest PG score (97\%), performing significantly better than LIME (76.85\%) and RISE (66.95\%).

\begin{figure*}[h!]
    \centering
    \includegraphics[width=0.19\linewidth]{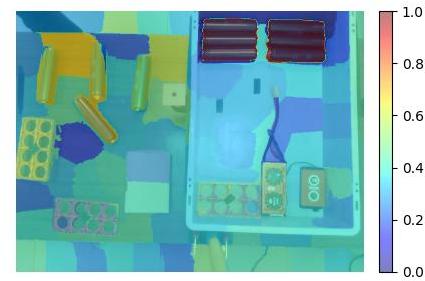}
    \includegraphics[width=0.19\linewidth]{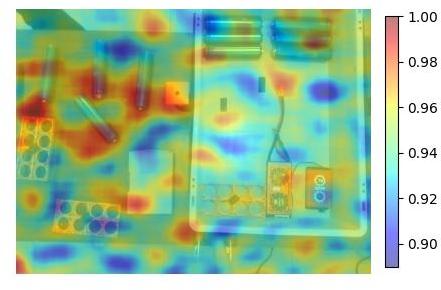}
    \includegraphics[width=0.19\linewidth]{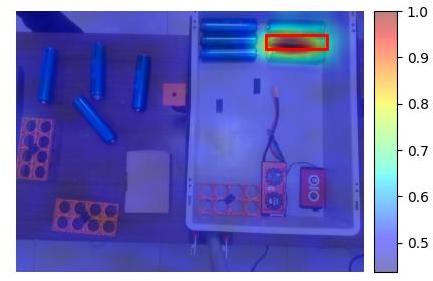}
    \includegraphics[width=0.19\linewidth]{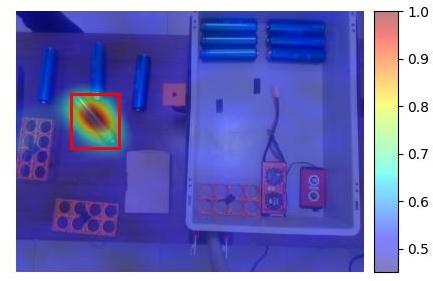}
    \includegraphics[width=0.19\linewidth]{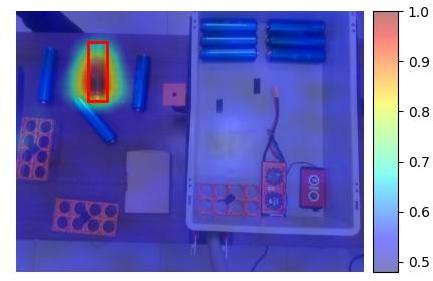}

    \includegraphics[width=0.19\linewidth]{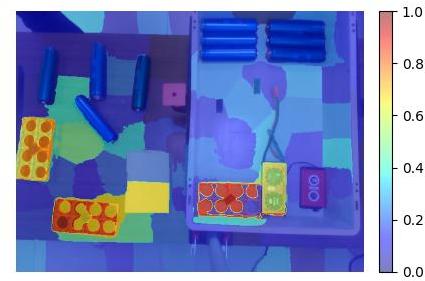}
    \includegraphics[width=0.19\linewidth]{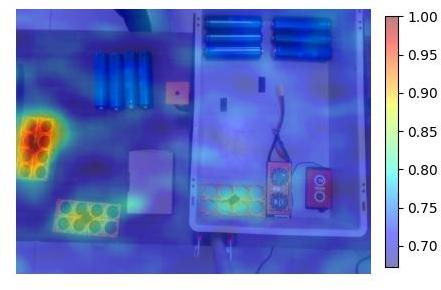}
    \includegraphics[width=0.19\linewidth]{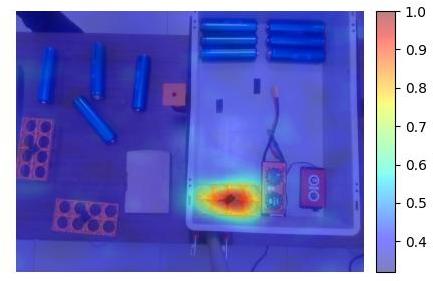}
    \includegraphics[width=0.19\linewidth]{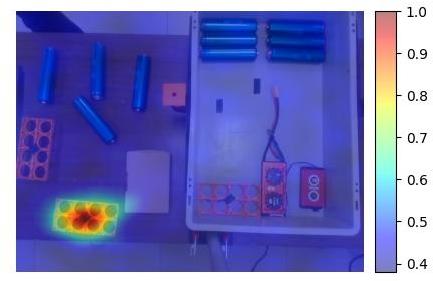}
    \includegraphics[width=0.19\linewidth]{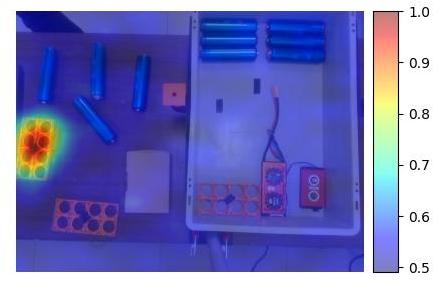}
    \caption{Heatmaps generated in a scene of the Battery Assembly dataset for two target classes: \texttt{individual battery} (top row) and \texttt{battery holder} (bottom row). The first and second columns shows the saliency maps obtained using LIME and RISE, independent of the number of objects of the same class in the image. Columns 3,4 and 5 display heatmaps generated using D-RISE for three different individual elements of the same class.}
    \label{fig:rob_heatmaps_lime_rise_drise}
\end{figure*}

Overall, when dealing with datasets containing only one object per class, the differences between LIME, RISE, and D-RISE are relatively small in quantitative terms. However, when multiple objects of the same class appear in a given input image, D-RISE clearly dominates over the rest of techniques. As illustrated in Figure \ref{fig:rob_heatmaps_lime_rise_drise}, D-RISE generates coherent heatmaps for each detected object in the Battery Assembly dataset, whereas LIME and RISE provide a global saliency map for the entire class. By combining the individual saliency maps from D-RISE, a more accurate and object-specific explanation can be produced. This also highlights the limitations of LIME and RISE when applied to multiple objects, as their global saliency maps do not differentiate between individual instances.

\begin{table*}[t]
    \centering
    \caption{Quantitative metrics of LIME, RISE and D-RISE for the Battery Assembly dataset. The table presents the performance of each XAI technique in terms of classification (Deletion, D-Deletion, Min-Subset, D-Min-Subset) and localization metrics (PG and EBPG), with scores reported for each object (\texttt{indiv batt, bms a, bms b, unknown object, batt holder}) and the overall average. 
    Lower values are better for metrics marked with $\downarrow$, while higher values are better for those marked with $\uparrow$. \textbf{Bold values} indicate the best average scores across all objects, highlighting the best-performing XAI method for each metric. Gray-highlighted values represent the best scores for each object category (\texttt{indiv batt, bms a, bms b, unknown object} or \texttt{batt holder}) and should be interpreted vertically.
    \vspace{2mm}}

    \resizebox{2\columnwidth}{!}{
        \begin{tabular}{cccccccc}
        \toprule
        \textbf{XAI Method} & \textbf{Object} & \textbf{Deletion} ($\downarrow$) & \textbf{D-Deletion} ($\downarrow$) & \textbf{Min-Subset} ($\downarrow$)& \textbf{D-Min-Subset} ($\downarrow$)& \textbf{PG} ($\uparrow$) & \textbf{EBPG} ($\uparrow$) \\
        \midrule
        
        \multirow{6}{*}{LIME} 
        & \texttt{indiv batt} & 0.7549 & 0.2806 & 44.3412 & 14.6756 & 0.3188 & \cellcolor[gray]{0.9} 2.5347 \\
        & \texttt{bms a} & 0.0184 & \cellcolor[gray]{0.9} 0.0181 & \cellcolor[gray]{0.9} 0.8784 & \cellcolor[gray]{0.9} 0.8784 & \cellcolor[gray]{0.9} 1.0000 & \cellcolor[gray]{0.9} 30.7573 \\
        & \texttt{bms b} & \cellcolor[gray]{0.9} 0.0125 & 0.0125 & \cellcolor[gray]{0.9} 0.8321 & \cellcolor[gray]{0.9}  0.8321 & \cellcolor[gray]{0.9} 1.0000 & \cellcolor[gray]{0.9} 16.5440 \\
        & \texttt{unknown object} & \cellcolor[gray]{0.9} 0.0624 & \cellcolor[gray]{0.9} 0.0245 &  \cellcolor[gray]{0.9} 2.0342 & \cellcolor[gray]{0.9}  2.0342 & \cellcolor[gray]{0.9} 1.0000 & \cellcolor[gray]{0.9} 20.1071 \\
        & \texttt{batt holder} & 0.1498 & 0.0849 & \cellcolor[gray]{0.9} 6.8115 & \cellcolor[gray]{0.9} 4.3766 & 0.5238 & \cellcolor[gray]{0.9} 10.2087 \\
        \cmidrule(l){2-8}
        & {Average} & \textbf{0.1996} & 0.0841 & \textbf{10.9795} & 4.5594 & 0.7685 & \textbf{16.0304} \\
        \midrule
        \multirow{6}{*}{RISE} 
        & \texttt{indiv batt} & 0.7659 & 0.4359 & 81.2344 & 32.0482 & 0.0144 & 1.2558 \\
        & \texttt{bms a} & 0.0190 & 0.0190 & 1.9417 & 1.9417 & \cellcolor[gray]{0.9} 1.0000 & 3.0409 \\
        & \texttt{bms b} & 0.0217 & 0.0217 & 2.1266 & 2.1266 & \cellcolor[gray]{0.9} 1.0000 & 2.4561 \\
        & \texttt{unknown object} & 0.5333 & 0.2008 & 3.8372 & 3.8372 & \cellcolor[gray]{0.9} 1.0000 & 5.5780 \\
        & \texttt{batt holder} & \cellcolor[gray]{0.9} 0.0902 & \cellcolor[gray]{0.9} 0.0595 & 7.9519 & 5.5632 & 0.3333 & 3.3681 \\
        \cmidrule(l){2-8}
        & {Average} & 0.2860 & 0.1474 & 19.4184 & 9.1034 & 0.6695 & 3.1398 \\
        \midrule
        \multirow{6}{*}{D-RISE} 
        & \texttt{indiv batt} & \cellcolor[gray]{0.9} 0.6214 & \cellcolor[gray]{0.9} 0.0311 & \cellcolor[gray]{0.9}  35.7678 & \cellcolor[gray]{0.9} 2.7725 & \cellcolor[gray]{0.9} 1.0000 & 2.0546 \\
        & \texttt{bms a}             & \cellcolor[gray]{0.9} 0.0181 & \cellcolor[gray]{0.9} 0.0181 &  1.9880 & 1.9880 & \cellcolor[gray]{0.9}  1.0000 & 3.2955 \\
        & \texttt{bms b}             & 0.0128 & \cellcolor[gray]{0.9} 0.0116 &  1.3407 & 1.3407 & \cellcolor[gray]{0.9} 1.0000 & 2.8876 \\
        & \texttt{unknown object}    & 0.0879 & 0.0485 &  3.7448 & 4.2071 & 0.8571 & 7.6831 \\
        & \texttt{batt holder}       & 0.4839 & 0.0626 & 21.7291 & 5.1009 & \cellcolor[gray]{0.9} 1.0000 & 4.9635 \\
        \cmidrule(l){2-8}
        & {Average}  & 0.2448 & \textbf{0.0344} & 12.9141 & \textbf{3.0819} & \textbf{0.9714} & 4.1768 \\
        \bottomrule
        \end{tabular}
    }
    \label{tab:rob_rq1}
\end{table*}

\subsection*{RQ2: D-Deletion metric for scenes with multiple objects of the same class}

As a secondary observation in the experiments of RQ1, the D-Deletion metric is specifically designed to overcome the limitations of traditional deletion metrics, particularly when multiple objects of the same class are present in an image.

In the Battery Assembly dataset, where several instances of the same class (e.g., \texttt{indiv batt}) appear, D-Deletion demonstrates clear advantages. By inspecting Table \ref{tab:rob_rq1}, RISE, while performing reasonably well with an average Deletion score of $0.2860$, it still obtains a relatively high D-Deletion score of 0.1474, suggesting that it struggles to differentiate between the contributions of individual objects. In contrast, D-RISE, which obtains an average Deletion score of 0.2448, outperforms RISE with a D-Deletion score of 0.03444. This highlights D-RISE’s ability to isolate and preserve key features for each object, providing more trustworthy, object-specific explanations rather than broad, class-level insights.

The Min-Subset and D-Min-Subset metrics, which measure the minimal proportion of pixels needed to disrupt a detection, reinforce these findings. In the Human-Robot dataset, Table \ref{tab:piap_rq1}, where only one object per class appears, the differences between Deletion and D-Deletion scores are minor, and the Min- Subset and D-Min-Subset values are close to each other. However, in the Battery Assembly dataset, where the differences between Deletion and D-Deletion are more substantial and multiple objects of the same class co-occur in the same image, the Min-Subset ($12.9141$) and D-Min-Subset ($3.0819$) values also diverge significantly.

\begin{figure*}
    \centering
    
    \begin{tabular}{llcccc}
    Class & XAI & $N=500$ & $N=1,000$ & $N=5,000$ & $N=10,000$ 
    \\
    \toprule
    \multirow{2}{*}{\rotatebox{90}{\texttt{Human}}} &
    \rotatebox{90}{D-RISE} &
    \includegraphics[width=0.2\linewidth]{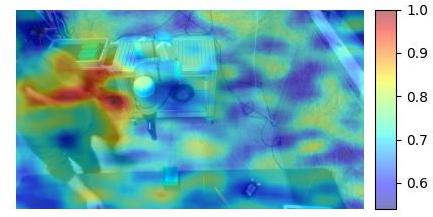} &
    \includegraphics[width=0.2\linewidth]{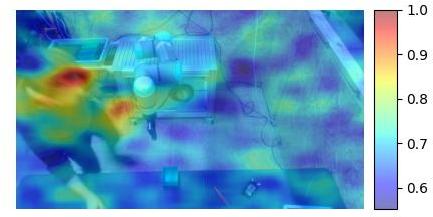} &
    \includegraphics[width=0.2\linewidth]{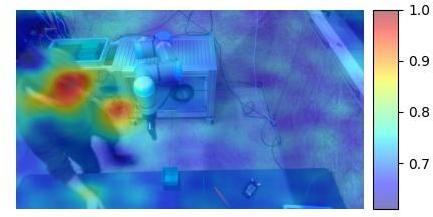} &
    \includegraphics[width=0.2\linewidth]{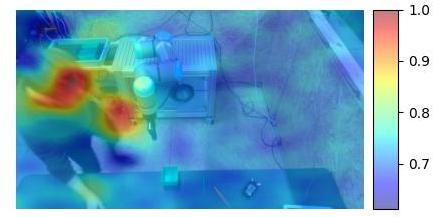}
    \\
    &
    \rotatebox{90}{D-MFPP} &
    \includegraphics[width=0.2\linewidth]{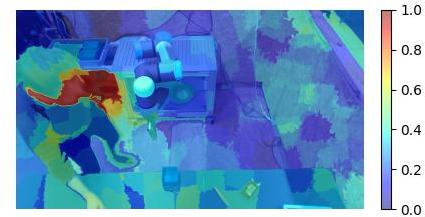} &
    \includegraphics[width=0.2\linewidth]{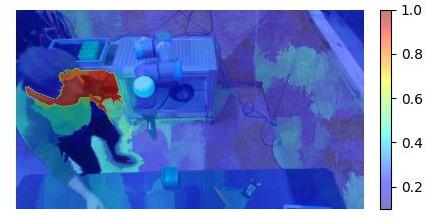} &
    \includegraphics[width=0.2\linewidth]{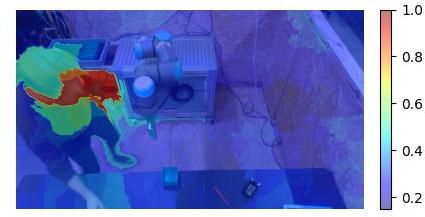} &
    \includegraphics[width=0.2\linewidth]{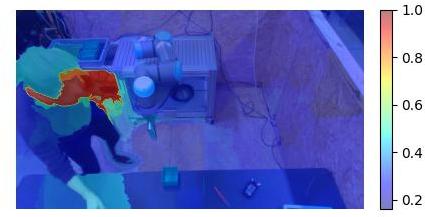}
    \\
    \cmidrule(l){2-6}

    \multirow{2}{*}{\rotatebox{90}{\texttt{Gripper}}} &
    \rotatebox{90}{D-RISE} &
    \includegraphics[width=0.2\linewidth]{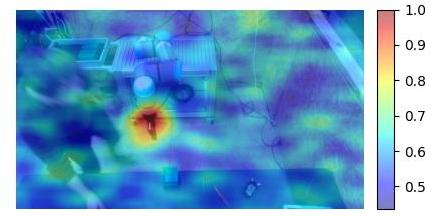} &
    \includegraphics[width=0.2\linewidth]{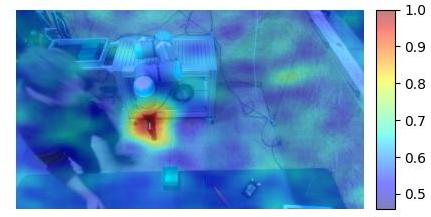} &
    \includegraphics[width=0.2\linewidth]{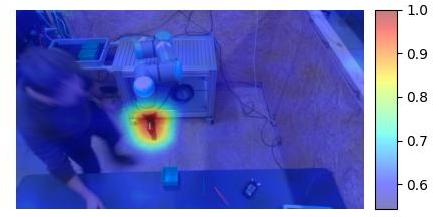} &
    \includegraphics[width=0.2\linewidth]{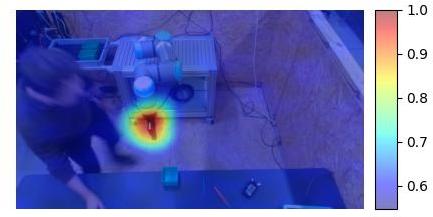}
    \\
    &
    \rotatebox{90}{D-MFPP} &
    \includegraphics[width=0.2\linewidth]{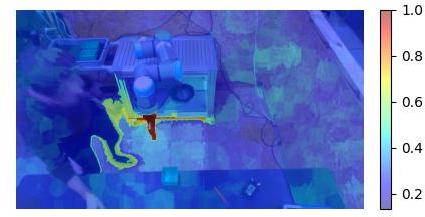} &
    \includegraphics[width=0.2\linewidth]{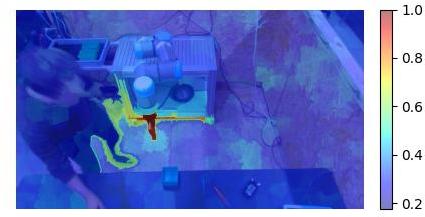} &
    \includegraphics[width=0.2\linewidth]{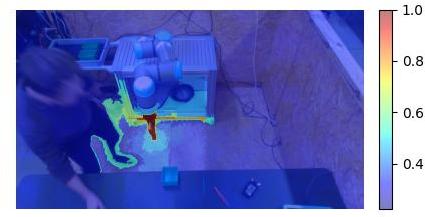} &
    \includegraphics[width=0.2\linewidth]{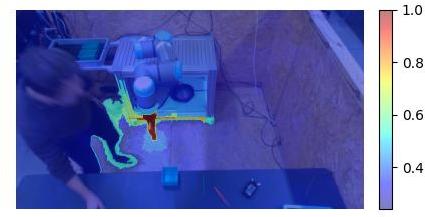}
    \\
    \midrule

    \multirow{2}{*}{\rotatebox{90}{\texttt{indiv batt}}} &
    \rotatebox{90}{D-RISE} &
    \includegraphics[width=0.2\linewidth]{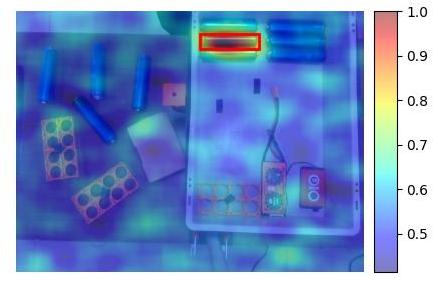} &
    \includegraphics[width=0.2\linewidth]{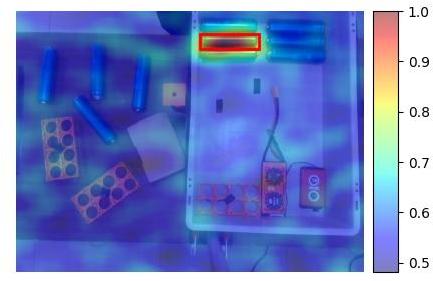} &
    \includegraphics[width=0.2\linewidth]{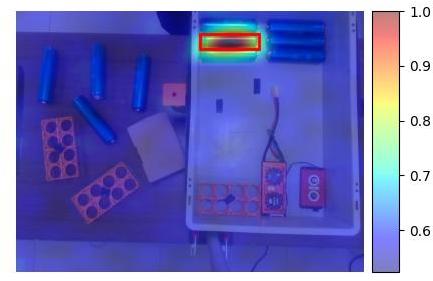} &
    \includegraphics[width=0.2\linewidth]{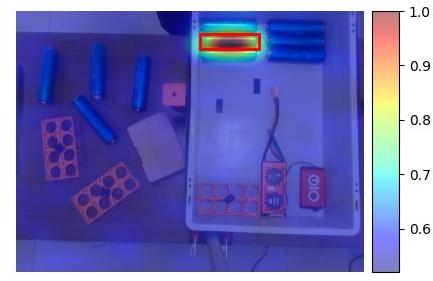}
    \\
    &
    \rotatebox{90}{D-MFPP} &
    \includegraphics[width=0.2\linewidth]{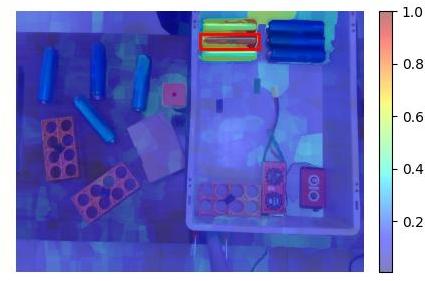} &
    \includegraphics[width=0.2\linewidth]{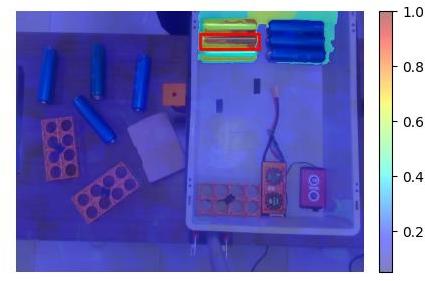} &
    \includegraphics[width=0.2\linewidth]{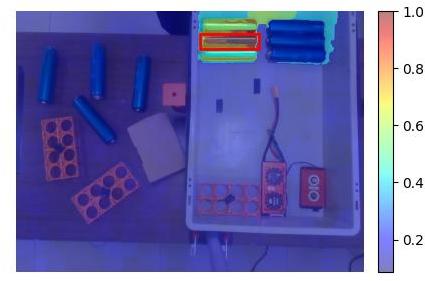} &
    \includegraphics[width=0.2\linewidth]{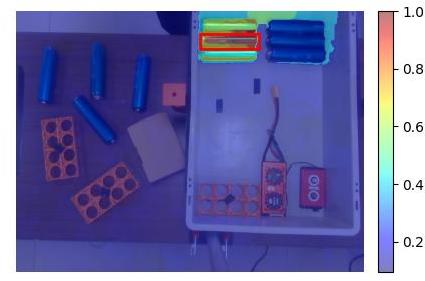}
    \\
    \cmidrule(l){2-6}

    \multirow{2}{*}{\rotatebox{90}{\texttt{bms a}}} &
    \rotatebox{90}{D-RISE} &
    \includegraphics[width=0.2\linewidth]{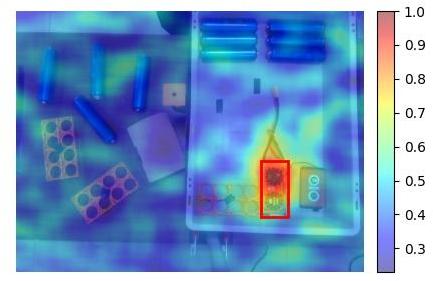} &
    \includegraphics[width=0.2\linewidth]{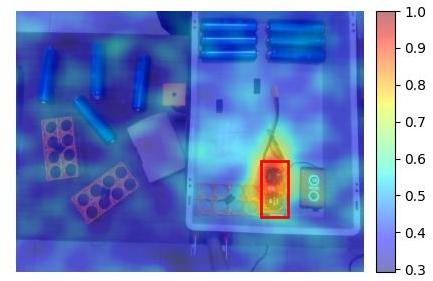} &
    \includegraphics[width=0.2\linewidth]{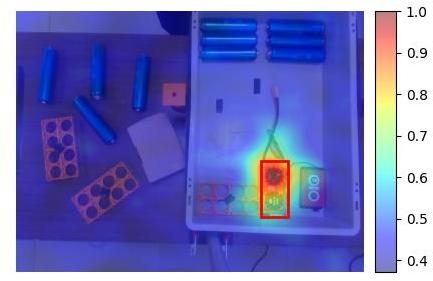} &
    \includegraphics[width=0.2\linewidth]{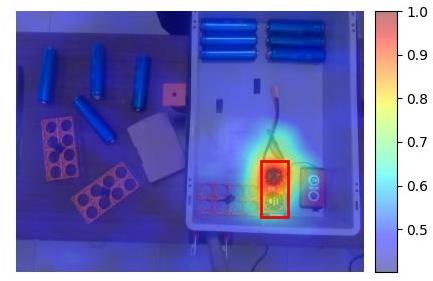}
    \\
    &
    \rotatebox{90}{D-MFPP} &
    \includegraphics[width=0.2\linewidth]{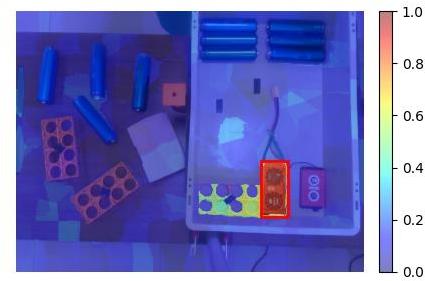} &
    \includegraphics[width=0.2\linewidth]{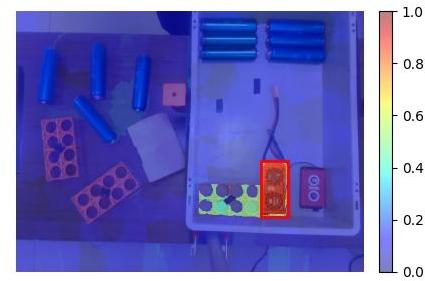} &
    \includegraphics[width=0.2\linewidth]{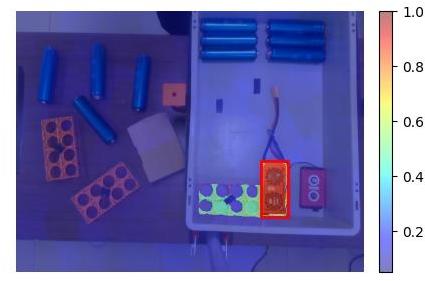} &
    \includegraphics[width=0.2\linewidth]{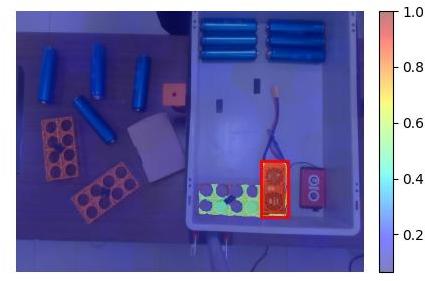}
    \\
    \bottomrule
    \end{tabular}
    
    \caption{Heatmaps obtained with D-RISE and D-MFPP using varying numbers of masks (500, 1000, 5000, 10000). Saliency maps are presented for two scenes, one from each dataset, demonstrating the evolution of heatmaps for a single element as the number of masks increases, highlighting the effect on explanation quality and focus.}
    \label{fig:num_masks_impact}
\end{figure*}

\begin{table*}[h]
    \centering
    \caption{Quantitative results for saliency maps generated with various masks techniques (D-Sliding Window, D-RISE, and D-MFPP) in the Human-Robot dataset. Only the average score across all classes is presented. 
    Results are interpreted vertically within each column (each metric represents a separate column).
    The best mask configuration for each XAI method and for each metric is highlighted in gray, while the overall best-performing XAI method with its optimal mask configuration is shown in bold.
    .\vspace{2mm}}
    \resizebox{2\columnwidth}{!}{
        \begin{tabular}{cccccccc}
        \toprule
        \textbf{XAI Method} & \textbf{Num Masks}  & \textbf{Deletion} ($\downarrow$) & \textbf{D-Deletion} ($\downarrow$) & \textbf{Min-Subset} ($\downarrow$)& \textbf{D-Min-Subset} ($\downarrow$)& \textbf{PG} ($\uparrow$) & \textbf{EBPG} ($\uparrow$) \\
        \midrule        
        \multirow{4}{*}{\makecell{D-Sliding Window\\(*)}} 
        & $w=32$, $s=8$ & 0.2323 & 0.2323 & 24.1966 & 24.1966 & \cellcolor[gray]{0.9} \textbf{1.0000} & 97.7128 \\ 
        & $w=32$, $s=16$ & 0.3009 & 0.3009 & 34.3990 & 34.3990 & \cellcolor[gray]{0.9} \textbf{1.0000} & 96.5487 \\ 
        & $w=64$, $s=8$ & 0.2046 & 0.2002 & 20.7980 & 20.7980 & \cellcolor[gray]{0.9} \textbf{1.0000} & 83.9141 \\
        & $w=64$, $s=16$ & \cellcolor[gray]{0.9} 0.2039 & \cellcolor[gray]{0.9} 0.1995 & \cellcolor[gray]{0.9} 20.7362 & \cellcolor[gray]{0.9} 20.7362 & \cellcolor[gray]{0.9} \textbf{1.0000} & \cellcolor[gray]{0.9} \textbf{83.5582} \\
        \midrule
        \multirow{4}{*}{D-RISE} 
        & 500 & 0.2645 & 0.0702 & 4.3520 & 4.3520 & 0.9062 & 10.7117 \\
        & 1000 & 0.2530 & 0.0519 & 4.2019 & 4.2019 & \cellcolor[gray]{0.9} 0.9375 & \cellcolor[gray]{0.9} 10.8382 \\
        & 5000 & \cellcolor[gray]{0.9} \textbf{0.2016} & 0.0438 & \cellcolor[gray]{0.9} 3.1713 & \cellcolor[gray]{0.9} 3.1713 & \cellcolor[gray]{0.9} 0.9375 & 10.7438 \\
        & 10000 & 0.2246 & \cellcolor[gray]{0.9} \textbf{0.0389} & 3.2496 & 3.2496 & \cellcolor[gray]{0.9} 0.9375 & 10.7094 \\
        \midrule
        \multirow{4}{*}{D-MFPP} 
        & 500 & 0.4137 & 0.0505 & \cellcolor[gray]{0.9} \textbf{2.9608} & \cellcolor[gray]{0.9} \textbf{2.9519} & \cellcolor[gray]{0.9} \textbf{1.0000} & 15.9700 \\
        & 1000 & 0.4227 & 0.0678 & 5.8648 & 5.8648 & \cellcolor[gray]{0.9} \textbf{1.0000} & 15.1166 \\
        & 5000 & 0.3944 & \cellcolor[gray]{0.9} 0.0469 & 3.8146 & 3.8058 & \cellcolor[gray]{0.9} \textbf{1.0000} & 16.0809 \\
        & 10000 & \cellcolor[gray]{0.9} 0.3893 & 0.0471 & 3.6734 & 3.6734 & \cellcolor[gray]{0.9} \textbf{1.0000} & \cellcolor[gray]{0.9} 16.2579 \\
        \bottomrule
        \multicolumn{8}{l}{*: D-Sliding Window scores have been computed excluding the cases where no explanations are produced for images with large-sized object instances.}\end{tabular}
        \label{tab:slidingw_rise_mfpp_piap}
    }
\end{table*} 
\begin{table*}[h]
    \centering
    \caption{Quantitative results for saliency maps generated with various masks techniques (D-Sliding Window, D-RISE, and D-MFPP) in the Battery Assembly dataset. 
    Results are interpreted vertically within each column (each metric represents a separate column).
    The best mask configuration for each XAI method and for each metric is highlighted in gray, while the overall best-performing XAI method with its optimal mask configuration is shown in bold.
    \vspace{2mm}}
    \resizebox{2\columnwidth}{!}{
        \begin{tabular}{cccccccc}
        \toprule
        \textbf{XAI Method} & \textbf{Num Masks} & \textbf{Deletion} ($\downarrow$) & \textbf{D-Deletion} ($\downarrow$) & \textbf{Min-Subset} ($\downarrow$)& \textbf{D-Min-Subset} ($\downarrow$)& \textbf{PG} ($\uparrow$) & \textbf{EBPG} ($\uparrow$) \\
        \toprule
        \multirow{4}{*}{\makecell{D-Sliding Window\\(*)}}
        & $w=32$, $s=8$ & 0.5198 & 0.3258 & 53.9463 & 31.2219 & \cellcolor[gray]{0.9}0.9855 & 90.4736 \\ 
        & $w=32$, $s=16$ & 0.4700 & 0.3046 & 45.9925 & 24.7660 & 0.9788 & 84.5302 \\ 
        & $w=64$, $s=8$ & 0.2578 & \cellcolor[gray]{0.9} \textbf{0.0245} & 26.4318 & \cellcolor[gray]{0.9} \textbf{2.4062} & \cellcolor[gray]{0.9}0.9855 & 55.8151 \\
        & $w=64$, $s=16$ & \cellcolor[gray]{0.9} \textbf{0.2570} & 0.0246 & \cellcolor[gray]{0.9} 26.3537 & 2.4772 & 0.9884 & \cellcolor[gray]{0.9}\textbf{54.8283} \\
        \midrule
        \multirow{4}{*}{D-RISE} 
        & 500 & 0.4307 & 0.0783 & 29.7843 & 4.9001 & \cellcolor[gray]{0.9}\textbf{0.9971} & 3.8152 \\
        & 1000 & 0.3837 & 0.0701 & 26.4242 & 3.3443 & \cellcolor[gray]{0.9}\textbf{0.9971} & 3.8154 \\
        & 5000 & \cellcolor[gray]{0.9} 0.3645 & 0.0431 & 25.2766 & 3.1374 & \cellcolor[gray]{0.9}\textbf{0.9971} & 3.8015 \\
        & 10000 & 0.3707 & \cellcolor[gray]{0.9} 0.0342 & \cellcolor[gray]{0.9} \textbf{23.3216} & \cellcolor[gray]{0.9} 2.9789 & \cellcolor[gray]{0.9}\textbf{0.9971} & \cellcolor[gray]{0.9}3.7883 \\
        \midrule
        \multirow{4}{*}{D-MFPP} 
        & 500 & 0.3327 & 0.0708 & 30.7555 & 5.6594 & 0.9942 & 6.2320 \\
        & 1000 & 0.3375 & 0.0630 & 30.0871 & 4.8843 & \cellcolor[gray]{0.9}\textbf{0.9971} & 6.1938 \\
        & 5000 & \cellcolor[gray]{0.9} 0.3042 & 0.0427 & \cellcolor[gray]{0.9} 27.8461 & \cellcolor[gray]{0.9} 3.4328 & \cellcolor[gray]{0.9}\textbf{0.9971} & 6.1772 \\
        & 10000 & 0.3100 & \cellcolor[gray]{0.9} 0.0409 & 29.6047 & 3.8185 & \cellcolor[gray]{0.9}\textbf{0.9971} & \cellcolor[gray]{0.9} 6.1680 \\
        \bottomrule
        \multicolumn{8}{l}{*: D-Sliding Window scores have been computed excluding the cases where no explanations are produced for images with large-sized object instances.}
        \end{tabular}
        \label{tab:slidingw_rise_mfpp_rob}
    }
    
\end{table*}

\subsection*{RQ3: Influence of the Mask Generation Strategy}

When comparing XAI approaches for object detection tasks configured with different mask generation techniques, the results in Tables \ref{tab:slidingw_rise_mfpp_piap} and \ref{tab:slidingw_rise_mfpp_rob} initially suggest that D-Sliding Window performs the best in almost all metrics. However, as noted in the captions, this is only in cases where explanations were provided. For the Human-Robot dataset (Table \ref{tab:slidingw_rise_mfpp_piap}), regardless of the window size and stride, D-Sliding Window failed to provide explanations for larger objects, such as \texttt{humans}, and only provided meaningful explanations for smaller objects like \texttt{gripper} instances. Similarly, in the Battery Assembly dataset (Table \ref{tab:slidingw_rise_mfpp_rob}), D-Sliding Window struggled with large objects when using a smaller window size (w=32), which led to higher scores in classification metrics. Even with an increased window size (w=64), some objects were still not detected, making it difficult to fairly compare with D-RISE and D-MFPP, both of which provided explanations for all detection proposals.

In contrast, D-RISE consistently performs best across all detection proposals when explanations are provided, particularly in terms of the D-Deletion metric, with D-MFPP as a close second. D-MFPP manages to better distinguish important regions by de-emphasizing less relevant areas, as reflected in the PG and EBPG localization scores. Yet, due to the superpixel segmentation approach used by D-MFPP, some relevant features are grouped with less important ones (see Figure \ref{fig:num_masks_impact}), causing a slight drop in classification metrics. As a result, D-MFPP trades off classification precision for improved localization and focus. D-Sliding Window, as previously noted, performs well on smaller objects but struggles to generate reliable explanations for larger objects. This limitation can be mitigated by using larger window sizes, which occlude larger portions of the image, thereby increasing the likelihood of capturing explanations for bigger objects, while smaller strides help produce more finely detailed explanations, as seen in Figure \ref{fig:slidingwindow_size_stride}.

\paragraph{Influence of the number of masks} This secondary analysis within RQ3 confirms that the number of masks in use has a clear impact on the  performance of the XAI method. Generally, more masks result in better outcomes. Interestingly, D-MFPP initially outperforms D-RISE in some metrics when using fewer masks (500 or 1,000), achieving a perfect PG score of 100\% in the Human-Robot dataset and a D-Deletion score of 0.0505 with just 500 masks. However, as the number of masks increases, D-RISE slightly surpasses D-MFPP in classification-related metrics. This trend can be also noted in Figure \ref{fig:num_masks_impact}, where (1) increasing the number of masks improves the quality of the saliency maps, and (2) D-MFPP provides cleaner maps with a lower number of masks. 

\subsection*{RQ4: Image dimension variation and YOLOv8 complexity}

\begin{table*}[h]
    \centering
    \caption{Explanations comparison for three image dimensions ($720\times 1280$, $360\times 640$, and $180\times 320$) and varying resolutions ($16\times 16$, $8\times 8$, and $4\times 4$) in the Human-Robot dataset.}
    \vspace{3mm}
    \resizebox{2\columnwidth}{!}{
        \begin{tabular}{cccccccc}
        \toprule
        \textbf{Image dimensions} & \textbf{Resolution} & \textbf{Deletion} ($\downarrow$) & \textbf{D-Deletion} ($\downarrow$) & \textbf{Min-Subset} ($\downarrow$)& \textbf{D-Min-Subset} ($\downarrow$)& \textbf{PG} ($\uparrow$) & \textbf{EBPG} ($\uparrow$) \\
        \midrule
        \multirow{3}{*}{$720\times 1280$}
        & $16\times 16$ & 0.2016 & \cellcolor[gray]{0.9} 0.0438 & \cellcolor[gray]{0.9}3.1713 & \cellcolor[gray]{0.9}3.1713 & 0.9375 & 10.7438 \\
        & $8\times 8$ & 0.0851 & 0.0512 & 4.0050 & 4.0050 & \cellcolor[gray]{0.9}1.0000 & 11.8188 \\
        & $4\times 4$ & \cellcolor[gray]{0.9} 0.0541 & 0.0504 & 4.0684 & 4.0684 & 0.9166 & \cellcolor[gray]{0.9}12.4406 \\
        \midrule
        \multirow{3}{*}{$360\times 640$}
        & $16\times 16$ & 0.4331 & 0.1697 & 10.6146 & 9.8644 & 0.7647 & 9.8634 \\
        & $8\times 8$  & 0.1889 &\cellcolor[gray]{0.9} 0.1003 & \cellcolor[gray]{0.9}6.1750 & \cellcolor[gray]{0.9}6.1750 & 0.8823 & 11.0002 \\
        & $4\times 4$  & \cellcolor[gray]{0.9} 0.1053 & 0.1013 & 8.2321 & 8.2321 & \cellcolor[gray]{0.9}0.9166 & \cellcolor[gray]{0.9}12.4447 \\
        \midrule
        \multirow{3}{*}{$180\times 320$}
        & $16\times 16$ & 0.3668 & \cellcolor[gray]{0.9} 0.0756 & \cellcolor[gray]{0.9}7.2829 & \cellcolor[gray]{0.9}7.2829 & 0.8957 & 10.5611 \\
        & $8\times 8$  & 0.2290 & 0.1206 & 10.1774 & 10.1774 & \cellcolor[gray]{0.9}1.000 & 11.6374 \\
        & $4\times 4$ & \cellcolor[gray]{0.9} 0.1531 & 0.1232 & 11.6246 & 11.6246 & 0.9545 & \cellcolor[gray]{0.9}12.0230 \\
        \bottomrule
        \end{tabular}
        \label{tab:img_dim_variation}
    }
\end{table*}
\begin{table*}[h]
    \centering
    \caption{Impact of YOLOv8 model complexity (nano, small, medium, large) on the explanations for the same experimental setup.}
    \vspace{3mm}
    \resizebox{1.7\columnwidth}{!}{
        \begin{tabular}{cccccccc}
        \toprule
        \textbf{Model size} & \textbf{Deletion} ($\downarrow$) & \textbf{D-Deletion} ($\downarrow$) & \textbf{Min-Subset} ($\downarrow$)& \textbf{D-Min-Subset} ($\downarrow$)& \textbf{PG} ($\uparrow$) & \textbf{EBPG} ($\uparrow$) \\
        \midrule
        Large & 0.1735 & 0.1569 & 7.1964 & 7.1964 & 0.8333 & 10.0317 \\
        Medium & 0.2016 & \cellcolor[gray]{0.9}0.0438 & \cellcolor[gray]{0.9}3.1713 & \cellcolor[gray]{0.9}3.1713 & \cellcolor[gray]{0.9}0.9375 & \cellcolor[gray]{0.9}10.7438 \\
        Small & 0.3120 & 0.2294 & 5.3750 & 4.3060 & 0.7352 & 9.6248 \\
        Nano  & \cellcolor[gray]{0.9}0.1434 & 0.0930 & 4.1177 & 4.1177 & 0.9117 & 10.2061 \\
        \bottomrule
        \end{tabular}
        \label{tab:model_dim_variation}
    }
\end{table*}

We further analyze the importance of image dimensions in the quality of explanations. For this purpose, we interpolate the image dimensions and train a new object detector for those image dimensions. As reported in Table \ref{tab:img_dim_variation}, larger image dimensions ($720\times 1280$) yield better results in terms of faithfulness metrics, degrading their value when decreasing the image resolution to $360\times 640$ and $180\times 320$. Nonetheless, PG and EBPG remain relatively stable across resolutions, suggesting that localization is less affected by image size than classification accuracy.

\begin{figure}[h!]
    \centering
    
    \begin{tabular}{cccc}
    \toprule
     & Class & Resolution$=4\times4$ & Resolution$=16\times16$ 
    \\
    \midrule
    \multirow{2}{*}{\rotatebox{90}{$720 \times 1280$}} &
    \rotatebox{90}{\texttt{Human}} &
    \includegraphics[width=0.3\columnwidth]{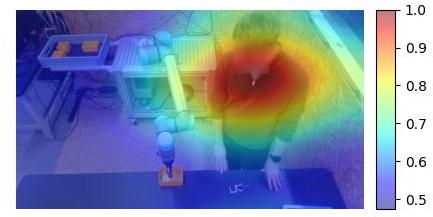} &
    \includegraphics[width=0.3\columnwidth]{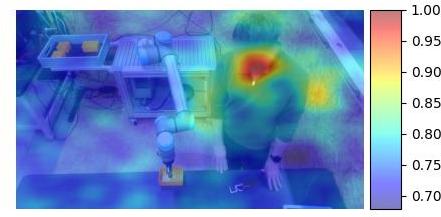}
    \\
    &
    \rotatebox{90}{\texttt{Gripper}} &
    \includegraphics[width=0.3\columnwidth]{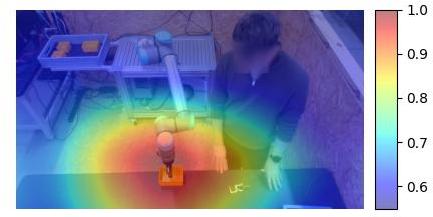} &
    \includegraphics[width=0.3\columnwidth]{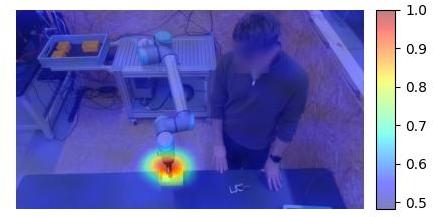}
    \\
    \midrule
    \multirow{2}{*}{\rotatebox{90}{$320 \times 640$}} &
    \rotatebox{90}{\texttt{Human}} &
    \includegraphics[width=0.3\columnwidth]{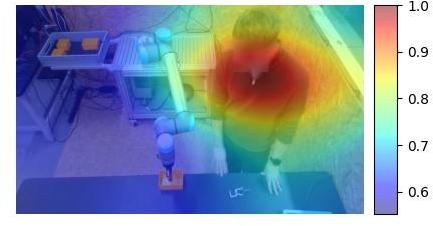} &
    \includegraphics[width=0.3\columnwidth]{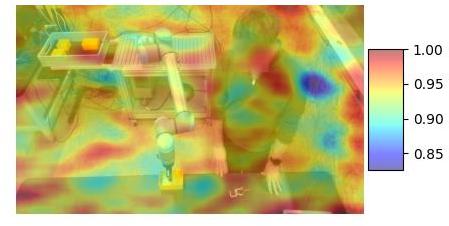}
    \\
    &
    \rotatebox{90}{\texttt{Gripper}} &
    \includegraphics[width=0.3\columnwidth]{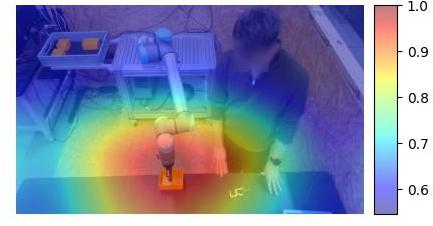} &
    \includegraphics[width=0.3\columnwidth]{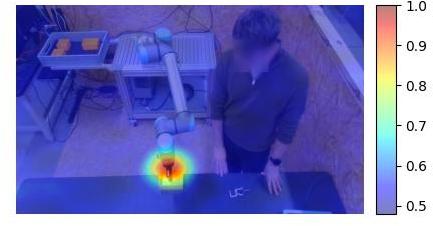}
    \\
    \midrule    
    \multirow{2}{*}{\rotatebox{90}{$180 \times 320$}} &
    \rotatebox{90}{\texttt{Human}} &
    \includegraphics[width=0.3\columnwidth]{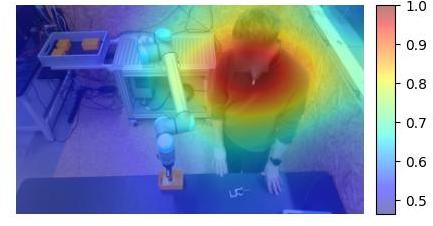} &
    \includegraphics[width=0.3\columnwidth]{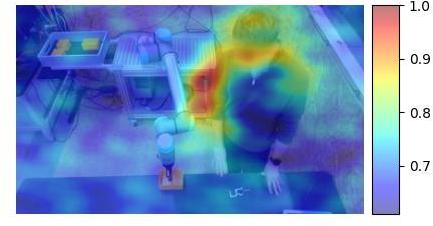}
    \\
    &
    \rotatebox{90}{\texttt{Gripper}} &
    \includegraphics[width=0.3\columnwidth]{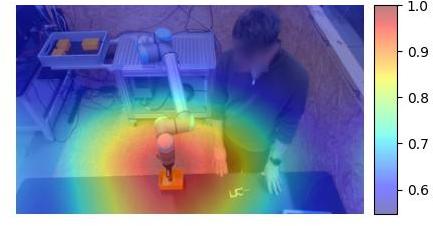} &
    \includegraphics[width=0.3\columnwidth]{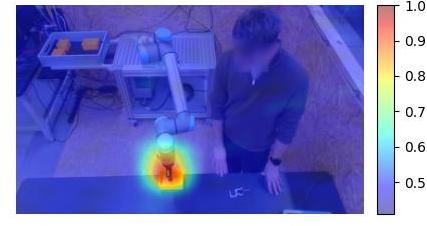}
    \\
    \bottomrule
    \end{tabular}
    
    \caption{Heatmaps generated with D-RISE ($p=0.25$ and resolutions of $4\times4$ and $16\times16$) on a scene from the Human-Robot scenario. The effect of varying input image dimensions can be observed.}
    \label{fig:img_dim_variation}
\end{figure}

Additionally, Table \ref{tab:img_dim_variation} shows the effect of applying D-RISE with masks generated at varying resolutions. While no clear trend emerges in the classification metrics (with some instances of Deletion increasing while D-Deletion decreases), the localization metrics generally improve with lower mask resolutions. This result is expected, as lower-resolution masks produce less fine-grained explanations but tend to highlight a larger area around the target object. This pattern is clearly visualized in Figure \ref{fig:img_dim_variation}.

Lastly, we investigate whether the use of different levels of parametric complexity of the YOLOv8 object detector affect the expected outcomes. When examining the results in Table \ref{tab:model_dim_variation}, no clear insights can be drawn, as the best results were reported by the Medium and Nano models. Nonetheless, it is important to highlight the performance of the Nano model, which despite requiring half the inference time and an order of magnitude less memory for deployment, achieves competitive results with respect to the rest of counterparts in the table. 


\section{Conclusion}\label{sec:conclusions}

This work has explored the performance of various XAI methods for object detection models, focusing on perturbation-based black-box explanation techniques like LIME, RISE, D-RISE applied to YOLOv8 in real-world object detection tasks in industrial setups. Our contribution is twofold: i) we have introduced D-MFPP, a method tailored for object detection that generates masks based on multi-level superpixels; and ii) we have proposed D-Deletion, a new quantitative metric that accounts for both class probability and localization when evaluating explanations.

Our results have demonstrated that D-RISE consistently outperforms LIME and RISE, particularly with the D-Deletion metric, which improves trustworthiness by delivering more focused, object-specific explanations. The mask generation process has been proven to play a key role in the quality of explanations, with more masks producing better explanations. The proposed D-MFPP technique has shown competitive performance with fewer masks, proving to be an efficient option in resource-constrained deployments. We have also found out that larger image dimensions improve classification metrics, while localization metrics remained stable across resolutions, suggesting that image size mainly affects classification. No clear patterns have emerged regarding model complexity; nevertheless, the \emph{nano} YOLOv8 model, with its low computational and memory demands, showed promise for real-time industrial applications.

\paragraph{Limitations of the study}While this study provides valuable insights into the explainability of object detectors in real-world scenarios, several limitations must be acknowledged:
\begin{itemize}[leftmargin=*]
    \item \textit{Limited data}: The real-world datasets considered in the study, particularly the Battery Assembly use case, are limited in size (only 7 data instances). This may restrict the generalization of our findings. Future studies with larger datasets embracing the methodology herein followed could provide more robust conclusions.


    \item \textit{Generalization of the object detector}: Although the methods presented here are model-agnostic, the study primarily focuses on YOLOv8 (i.e., one stage detector). Extending this work to consider other object detector architectures, such as Faster R-CNN (i.e., two stage detector) or Detection Transformers (DETR, RT-DETR) could ease a broader understanding of XAI methods across different neural architectures for this modeling task.

    \item \textit{Computation time}: The computation time required to generate explanations can be significant, especially when using perturbation-based methods. Approaches that reduce the number of masks used, or optimize mask generation, could help mitigate this limitation and improve efficiency.
\end{itemize}

\paragraph{Future Research}We plan to explore hierarchical masking approaches \cite{yan_gsm-hm_2022,yan_model-agnostic_2024,moradi_model-agnostic_2024} and feature fusion techniques \cite{truong_towards_2024} that further refine pixel saliency across multiple levels, reducing masking efforts and enhancing the quality of the saliency maps. Moreover, we will explore techniques to reduce the computation time while maintaining performance and explanation quality \cite{nguyen_efficient_2024}, which can be crucial in real-world critical applications wherein explanations have to be furnished fastly and presented to the human for its supervision and validation.

\section*{Acknowledgments}
\begin{figure}[h!]
    \centering    \includegraphics[width=0.3\linewidth]{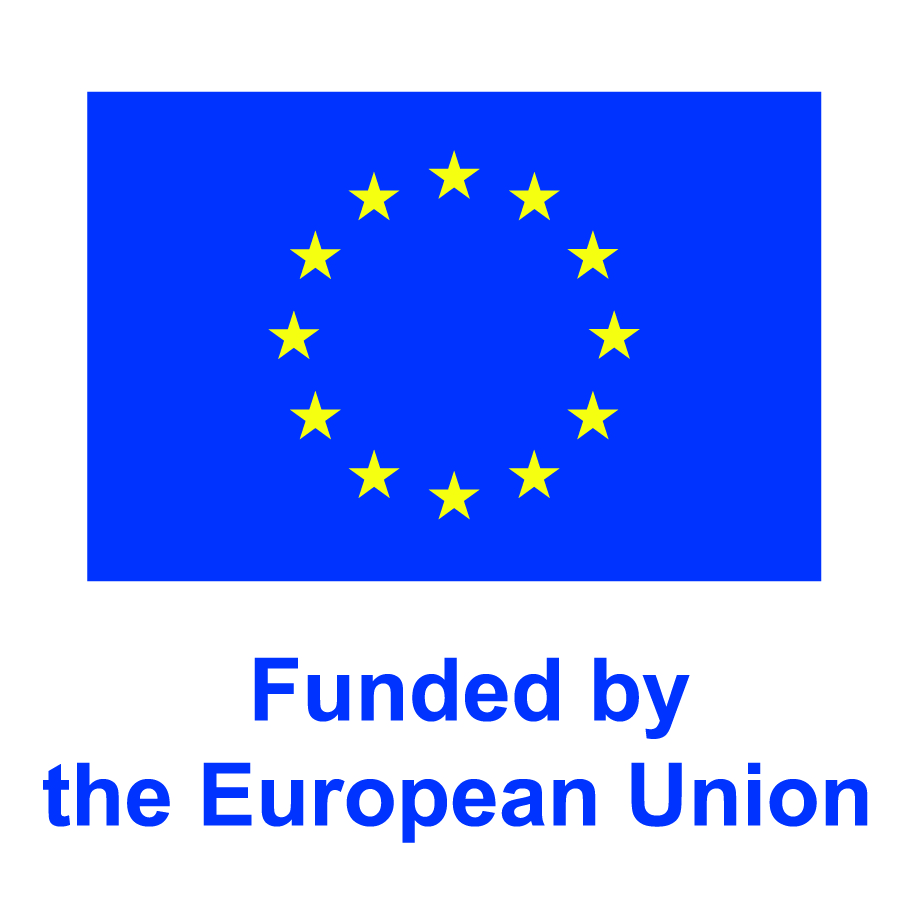}
    \includegraphics[width=0.3\linewidth]{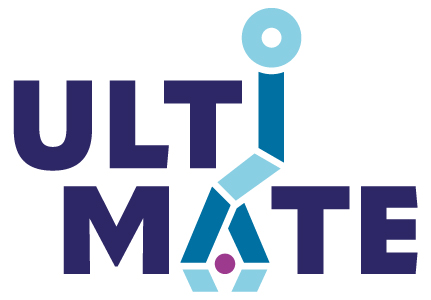}
\end{figure}

A. Andres, I. Laña and J. Del Ser receive support from the ULTIMATE project (ref.~101070162) funded by the European Commission under the HORIZON-CL4-DIS program (HORIZON-CL4-2021-HUMAN-01). J. Del Ser and I. Laña also acknowledge funding from the Basque Government (MATHMODE, IT1456-22).

\section*{Author Contributions}
\noindent \textbf{Alain Andres}: Conceptualization, Data curation, Formal analysis, Investigation, Methodology, Software, Writing (Original draft preparation). \textbf{Aitor Martinez-Seras}: Validation, Writing (Original draft preparation). \textbf{Ibai La\~na}: Supervision, Project administration, Writing (Review and Editing). \textbf{Javier Del Ser}: Conceptualization, Supervision, Validation, Writing (Reviewing and Editing).

\section*{Declaration of competing interest}
The authors declare that they have no conflicts of interest regarding this work. However, views and opinions expressed are those of the author(s) only and do not necessarily reflect those of the European Union. The European Union can not be held responsible for them.

\section*{Declaration of Generative AI and AI-assisted technologies in the writing process}
During the preparation of this work the author(s) used OpenAI's ChatGPT4 in order to enhance the clarity of the writing, improve the readability of the manuscript, and check for possible English language mistakes. After using this tool/service, the authors reviewed and edited the content as needed, and take full responsibility for the content of the publication.

\bibliographystyle{elsarticle-harv}
\bibliography{bibliography}

\end{document}